\documentclass{article}

\usepackage{graphicx} 
\usepackage{float}
\usepackage{authblk}

\usepackage[letterpaper,top=2cm,bottom=2cm,left=3cm,right=3cm,marginparwidth=1.75cm]{geometry}
\usepackage{algorithm}
\usepackage{algpseudocode}
\usepackage{amsmath}
\usepackage{multirow}
\usepackage{booktabs}
\usepackage{float}
\usepackage{makecell}
\usepackage{CJKutf8}
\usepackage[numbers,sort&compress]{natbib}
\usepackage{enumitem}

\usepackage[colorlinks=true, urlcolor=blue, linkcolor=red]{hyperref}

\title{Transcending Language Boundaries: Harnessing LLMs for Low-Resource Language Translation}

\newcommand{\equalcontrib}{\textsuperscript{*}}

\author[1]{Peng Shu\thanks{Co-first author}}
\author[1]{Junhao Chen\equalcontrib}
\author[1]{Zhengliang Liu}
\author[2]{Hui Wang}
\author[1]{Zihao Wu}
\author[4]{Tianyang Zhong}
\author[1]{Yiwei Li}
\author[1]{Huaqin Zhao}
\author[1]{Hanqi Jiang}
\author[1]{Yi Pan}
\author[1]{Yifan Zhou}
\author[3]{Constance Owl}
\author[5,6,7]{Xiaoming Zhai}
\author[1,5,6]{Ninghao Liu}
\author[3]{Claudio Saunt}
\author[1,5,6]{Tianming Liu\thanks{Corresponding author}}

\affil[1]{School of Computing, The University of Georgia, Athens 30602, USA}
\affil[2]{Second Language Acquisition and Teaching, University of Arizona, Tucson, 85721, USA}
\affil[3]{Department of History, The University of Georgia, Athens 30602, USA}
\affil[4]{Department of Mathematical and Statistical Sciences, University of Alberta, Edmonton, Canada }
\affil[5]{National GENIUS Center, Athens, GA, USA}
\affil[6]{AI4STEM Education Center, University of Georgia, Athens, GA, USA}
\affil[7]{Department of Mathematics, Science, and Social Studies Education, University of Georgia, Athens, GA, USA}

\begin{document}
\maketitle

\begin{abstract}
Large Language Models (LLMs) have demonstrated remarkable success across a wide range of tasks and domains. However, their performance in low-resource language translation, particularly when translating into these languages, remains underexplored. This gap poses significant challenges, as linguistic barriers hinder the cultural preservation and development of minority communities. To address this issue, this paper introduces a novel retrieval-based method that enhances translation quality for low-resource languages by focusing on key terms, which involves translating keywords and retrieving corresponding examples from existing data. To evaluate the effectiveness of this method, we conducted experiments translating from English into three low-resource languages: Cherokee, a critically endangered indigenous language of North America; Tibetan, a historically and culturally significant language in Asia; and Manchu, a language with few remaining speakers. Our comparison with the zero-shot performance of GPT-4o and LLaMA 3.1 405B, highlights the significant challenges these models face when translating into low-resource languages. In contrast, our retrieval-based method shows promise in improving both word-level accuracy and overall semantic understanding by leveraging existing resources more effectively. 
\end{abstract}

\section{Introduction}
Low-resource languages, often spoken by small and marginalized communities, face critical challenges in preservation, communication, and cultural transmission. Many of these languages lack extensive written documentation or digital resources, which hinders their use in critical domains like healthcare, education, and public services. The absence of adequate linguistic resources not only threatens the survival of these languages but also creates barriers to services, often leaving native speakers disconnected from modern applications and opportunities. In healthcare, for example, the inability to communicate in one's native language can lead to misunderstandings between patients and medical professionals, potentially impacting the quality of care and patient outcomes. The same challenges apply to legal, educational, and government services, where effective communication is essential. For these communities, access to reliable and accurate translations in their native language is crucial.

Addressing this problem with traditional methods of machine translation has proven challenging\cite{Liu2023}. Historically, machine translation has been dominated by well-resourced languages like English, French, and Chinese, with significant advancements in translation quality being driven by the availability of large, high-quality bilingual corpora. However, low-resource languages often lack parallel text corpora, limiting the effectiveness of rule-based and statistical approaches. As a result, machine translation for low-resource languages has lagged behind, exacerbating the digital divide for these communities.

In recent years, the development of large language  models (LLMs) has opened up new possibilities for low-resource language translation. Unlike traditional methods, LLMs benefit from vast pretraining across multiple languages, enabling them to generate coherent text even in languages for which little data exists. The transformer-based architecture that underpins these models, particularly models like GPT-4 \cite{achiam2023gpt} and LLaMA \cite{dubey2024llama}, has revolutionized the field of NLP by allowing parallel processing and improved contextual understanding. These models have demonstrated strong performance in many tasks, including translation, question-answering, and text generation, with emergent abilities that allow them to handle languages with scarce training data.

The evolution of natural language processing from early techniques such as word embeddings \cite{mikolov2013efficient} to current models highlights how far the field has come. The development of word embeddings in the early 2010s allowed words to be represented as dense vectors, capturing their semantic relationships. This led to the rise of sequence-to-sequence models \cite{sutskever2014sequence}, which transformed machine translation by enabling models to map input sequences (such as sentences in one language) to output sequences (sentences in another language) with a higher degree of fluency and accuracy. The introduction of the Transformer architecture \cite{vaswani2017attention} marked a turning point in the field, with its attention mechanisms enabling models to process entire sentences at once, rather than word by word, resulting in significant improvements in translation tasks.

The field of machine translation has evolved from rule-based systems to statistical machine translation (SMT) \cite{koehn2003statistical}, which relies on probabilistic models to generate translations based on statistical patterns in bilingual text. SMT systems dominated the field in the 1990s and early 2000s, but they required large volumes of parallel text to function effectively. Neural machine translation (NMT) \cite{sutskever2014sequence, bahdanau2014neural} soon emerged, leveraging deep learning and sequence-to-sequence models to generate more fluent and accurate translations. NMT marked a significant improvement over SMT by allowing for better context management across long sequences of text. Transformer-based models such as MarianMT \cite{junczys2018marian} and OpenNMT \cite{klein2017opennmt} had set impressive standards in machine translation, particularly for high-resource languages. Generative LLMs have further pushed the boundaries of flexibility and knowledge representation, surpassing previous models in their ability to handle diverse tasks with minimal task-specific training. LLMs like GPT-4 \cite{achiam2023gpt} and LLaMA \cite{dubey2024llama} leverage massive datasets across numerous languages, allowing them to generate human-like text across diverse domains and tasks, including translation.

Despite these advancements, generative LLMs still face significant challenges when applied to low-resource languages. One key issue is that these languages are often severely underrepresented in the pretraining data, leading to poor generalization when translating between high-resource and low-resource languages. Even with methods such as zero-shot and few-shot learning, generative LLMs may produce translations that are inaccurate or nonsensical due to limited exposure to the linguistic nuances of low-resource languages. Hallucination, where the model generates information that does not exist in the source text, is particularly problematic in this context, as it undermines the reliability of translations for critical domains like healthcare, education, and public services \cite{jayakody2024performance}.

One promising approach to improving translation for low-resource languages is Retrieval-Augmented Generation (RAG) \cite{fan2024survey}. By integrating external knowledge retrieval with language generation, RAG enhances the model’s ability to provide accurate and context-aware translations. This hybrid framework allows the model to fetch relevant data during the translation process, thereby addressing limitations such as outdated or incomplete internal knowledge. RAG can be particularly beneficial for low-resource languages, as it allows the model to compensate for the scarcity of training data by pulling information from related high-resource languages or domain-specific corpora.

In this paper, we explore the performance of two cutting-edge models, GPT-4o and LLaMA 3.1, in translating from a high-resource language (English) into Cherokee, a critically endangered Native American language. We evaluate these models using both automatic evaluation metrics, such as BLEU~\cite{papineni2002bleu}, ROUGE~\cite{lin2004rouge}, and BERTScore~\cite{zhang2019bertscore}, and human expert assessments to gauge the quality of the translations. Our goal is to assess how well these models generalize to languages with limited training data and to identify strategies, such as keyword-based retrieval, that can further improve translation quality for low-resource languages. 

In the broader context, improving LLM translation for low-resource languages can have far-reaching implications. It can not only help bridge the communication gap in critical domains like healthcare but also contribute to the preservation and revitalization of endangered languages. By making low-resource languages more accessible and usable in digital environments, we can help uplift communities and ensure that these languages continue to thrive in the modern world.

\section{Related Work}

\subsection{Large Language Models}

LLMs have revolutionized NLP, with applications across diverse fields such as education, healthcare, robotics, etc \cite{lee2023multimodality,shu2024llms,latif2023artificial,wang2024large,liu2024understanding,latif2024systematic,li2024artificial,zhong2024evaluation}. The foundation of LLMs lies in the transformer architecture introduced by Vaswani et al. \cite{vaswani2017attention}. With its self-attention mechanism, the transformer effectively handles long-range dependencies, surpassing earlier models like Recurrent Neural Networks (RNNs) \cite{cho2014learning} and Long Short-Term Memory (LSTM) networks \cite{graves2012long}, which struggled with processing long sequences and parallelization. Transformers have achieved state-of-the-art (SOTA) performance on tasks such as machine translation. The BERT model \cite{devlin2018bert}, developed from Transformers, further enhanced the efficiency and scalability of NLP models through self-supervised pretraining.

More recently, generative language models, particularly the GPT series \cite{radford2019language, brown2020language}, were the first to demonstrate zero-shot and few-shot in-context learning abilities without being finetuned on specific downstream tasks. These decoder-only models, pretrained on massive datasets from diverse sources, integrate a wide range of world knowledge, enabling efficient autoregressive generation with emergent abilities and scaling to larger and more powerful models. These factors are key to today's LLMs. InstructGPT \cite{ouyang2022training} refined the framework by incorporating reinforcement learning from human feedback (RLHF) \cite{christiano2017deep}, aligning model outputs more closely with human preferences.

Building on these advancements, OpenAI released ChatGPT in 2022, marking a new milestone in LLM development. Its large-scale pretraining and RLHF mechanisms enabled it to excel in tasks such as question answering, summarization, and dialogue, making it widely popular for its human-like conversational abilities. Subsequent models like GPT-4 and GPT-4V further integrated multimodal capabilities and enhanced reasoning abilities, enabling more complex tasks. Open-source LLMs such as LLaMA \cite{touvron2023llama} and Mistral \cite{jiang2023mistral} have also demonstrated competitive performance, pushing LLMs toward more open and widespread use in real-world applications. Indeed, LLMs and related vision-language models have demonstrated significant practical impact in diverse domains~\cite{liu2023summary,zhao2023brain,ma2024iterative,dai2023chataug,liu2023radiology,liao2023differentiating,liu2023context,rezayi2022clinicalradiobert,dai2023ad,zhao2023ophtha,zhang2024generalist,liu2023radonc,liu2024fine,lyu2024gp,wang2024comprehensive,huang2024position,liu2024surviving,huang2024trustllm,tian2024assessing}.

With these advancements, the integration of LLMs into various real-world applications has garnered increasing attention. LLMs' human-like understanding and reasoning abilities pave the way toward Artificial General Intelligence (AGI) and can facilitate societal development across a wide range of domains \cite{zhao2023brain,liu2023transformation,zhenyuan2024analyzing,zhao2024revolutionizing,li2024large,lee2023multimodality}.

\subsection{Machine Translation on Low-Resource Language}
Machine Translation for low-resource languages has been a long-standing challenge in the field of NLP. While machine translation for high-resource languages, such as English, Chinese, or Spanish, has seen considerable improvements, particularly with the advent of NMT techniques, low-resource languages have lagged due to the scarcity of large parallel corpora and linguistic resources.

Early efforts in machine translation, particularly for low-resource languages, were based on rule-based and SMT approaches. Rule-based systems relied on linguistic rules crafted by experts and extensive lexicons, but they were often labor-intensive and brittle when applied to complex languages. SMT, introduced in the 1990s, offered a data-driven approach, where models learned translation probabilities from bilingual text corpora. However, SMT required large volumes of parallel data, which was a significant limitation for low-resource languages, and its reliance on phrase-based techniques often failed to capture complex linguistic phenomena like morphology and syntax in underrepresented languages.

With the advent of Neural Machine Translation (NMT), particularly sequence-to-sequence (Seq2Seq) models with attention mechanisms, the field of Machine Translation entered a new era. NMT demonstrated superior performance over SMT in many language pairs by leveraging deep learning to create richer representations of source and target sentences. However, NMT's effectiveness heavily depends on the availability of large datasets, making it less applicable to low-resource settings. Researchers began exploring techniques to alleviate the data scarcity issue. Transfer learning became a popular method, where models pretrained on high-resource languages are fine-tuned on low-resource languages. This technique allowed low-resource languages to benefit from knowledge gained from related high-resource languages. Multilingual NMT further extended this idea, training models to translate between multiple languages simultaneously. This method improved performance for low-resource languages by allowing them to share representations with high-resource languages, leveraging multilingual data in a shared model architecture.

Data augmentation techniques, such as back-translation, have become widely used for low-resource Machine Translation. In back-translation, a model is used to translate monolingual target language data into the source language, thereby creating pseudo-parallel corpora. This approach significantly boosts training data for low-resource pairs and has been shown to improve translation quality in languages with limited direct parallel data. Additionally, unsupervised machine translation emerged as a promising approach for low-resource languages by eliminating the need for parallel data entirely. Unsupervised Machine Translation leverages monolingual corpora in both the source and target languages and learns to align and translate between them using iterative back-translation and shared latent representations. However, while these methods show promise, they still struggle with low-resource languages that have extremely limited or non-existent monolingual resources.

More recently, LLMs such as GPT-3, GPT-4, and LLaMA have emerged as powerful tools for a range of NLP tasks, including translation. These models, with billions of parameters and trained on extensive multilingual data, have shown promise in addressing low-resource translation through prompt-based methods. By employing prompt engineering, LLMs can perform zero-shot or few-shot translation tasks by conditioning on examples or instructions provided as input prompts. However, their performance on low-resource languages, particularly those with complex linguistic features or limited presence in training data (e.g., indigenous languages like Cherokee), remains an area of active research.

\section{Low-Resource Language}
\subsection{Cherokee}
In this paper, we focus on one Native American low-resource language: Cherokee. We also design a LLMs empowered Machine Translation task based on Cherokee.
The Cherokee language, known as Tsalagi gawonihisdi in its native form, belongs to the Iroquoian language family. It is primarily spoken by the Cherokee people, whose traditional homelands are in what is now the southeastern United States.  Today, as a result of the federal policy to deport Native peoples from their homes east of the Mississippi River in the 1830s, there are two Cherokee nations in Oklahoma (the Cherokee Nation and the United Keetoowah Band of Cherokee Indians) and one in North Carolina (the Eastern Band of Cherokee Indians). Despite centuries of colonial pressures, the Cherokee language has persisted through various revitalization efforts. However, it still remains classified as a critically endangered language according to UNESCO \cite{zhang2022can}.

The historical trajectory of the Cherokee language is inseparable from the history of the Cherokee people. Before European contact, Cherokee was an exclusively oral language \cite{zhang2020chren}. The Cherokee syllabary, created in the early 19th century by Sequoyah (also known as George Guess), revolutionized the language by providing a standardized writing system. The syllabary consists of 85 symbols, each representing a syllable, making it fundamentally different from an alphabetic system like English, where letters represent individual phonemes. Cherokee is a polysynthetic language, meaning that words often consist of multiple morphemes that combine to convey complex meanings. In Cherokee, single words can function as complete sentences in English. When discussing parts of speech, verbs are the most morphologically complex part of the Cherokee language. A typical Cherokee verb consists of several components:
\begin{itemize}
\item \textbf{Pronominal prefix}, which indicates the subject or object
\item \textbf{Prepronominal prefix}, that provides additional information such as tense, aspect, or negation
\item \textbf{Verb root}, the core meaning of the verb
\item \textbf{Aspect markers}, indicating whether an action is completed or ongoing
\item \textbf{Directional suffixes}, specifying the direction of the action
\end{itemize}
Nouns in Cherokee however, are typically simpler than verbs. They may consist of a root and optional suffixes. But they do not show the same level of inflection as verbs. Cherokee does not mark nouns for gender or case, although it does differentiate between animate and inanimate entities.
Cherokee syntax follows a verb-final word order (SOV). In other words, Cherokee sentences typically have the subject first, followed by the object, and then the verb. However, Cherokee exhibits considerable flexibility in its word order due to the rich morphological marking of verbs. It allows speakers to emphasize different parts of a sentence by rearranging the word order without losing clarity.

The Cherokee language is crucial for preserving cultural identity and heritage, as it carries traditional knowledge, oral stories, and ceremonies that define the Cherokee people. It has historical significance due to the creation of the Cherokee syllabary, which empowered the Cherokee Nation in the colonial era and remains a symbol of resilience. Cherokee contributes to linguistic diversity, offering insights into unique grammatical structures and language typology. It also hold valuable culture knowledge. Efforts to revitalize the language support indigenous sovereignty and ensure intergenerational communication, allowing future generations to retain their cultural roots. We hope by alleviating LLMs, linguists and computer science researchers can contribute to revitalize Cherokee. It is one of the most vital goals for this paper.
\subsection{Tibetan}
Tibetan, a language with a rich historical legacy, is first attested from the mid-seventh century CE during the Tibetan Empire, which adopted the Chinese style of documenting political events. The earliest form, known as Old Tibetan, is found in the empire's documents and inscriptions, while Classical Tibetan emerged after a significant language reform and standardization in the ninth century \cite{thurgood2016sino}.

Morphologically, Tibetan shows both simplicity and complexity. Many basic vocabulary items are monosyllabic, reflecting its Sino-Tibetan roots, but the language also contains a significant number of bimorphemic and disyllabic nouns. Compound nouns and light verb constructions (Noun + Verb combinations) are prevalent, and while they are common in modern Tibetan, they do not form compounds in the same formal sense as compound nouns \cite{beyer1992classical}.

Tibetan utilizes tones to differentiate words that are otherwise phonetically identical, although the presence and nature of tones vary by dialect. Classical Tibetan, from which modern Tibetan dialects evolved, was not tonal, and the development of tones is a relatively recent phenomenon in the language's evolution. Tibetan is primarily a Subject-Object-Verb (SOV) language. It uses postpositions rather than prepositions, and word order plays a critical role in sentence structure. The language has an ergative-absolutive alignment, which is evident in its case marking and verb agreement patterns.

Tibetan verbs are often complex and consist of several components, particularly in classical and literary forms. A typical Tibetan verb consists of the following components:

\begin{itemize}
\item \textbf{Pronominal prefix}, which indicates the person (subject or object) involved in the action
\item \textbf{Verb root}, the core lexical meaning of the verb
\item \textbf{Tense markers}, which indicate past, present, or future actions
\item \textbf{Auxiliary verbs}, which modify the meaning of the main verb (e.g., indicating ability or necessity)
\item \textbf{Honorific markers}, used to show respect, particularly in formal or religious contexts
\end{itemize}

These components allow Tibetan verbs to convey detailed grammatical and social information, making them highly versatile and integral to the language's structure.

\subsection{Manchu}
Manchu, part of the Tungusic language family, was historically the language of the ruling elite of the Qing Dynasty, which governed China from 1644 to 1912. The Manchu language, spoken by the Manchu people, was used extensively in the imperial court for religious ceremonies, diplomatic and ideological discourse, and for communicating with bannermen, nobility, and military officials \cite{Crossley1993}. Today, however, Manchu is considered a highly endangered language, with fewer than 100 native speakers left, predominantly among older generations in the remote regions of northeast China \cite{DeGruyter2006}. In Heilongjiang, the northernmost province of Manchuria, Manchu was still spoken by a considerable population at the start of the twentieth century. By the century’s end, however, only a few small groups of middle-aged and elderly speakers remained, and the outlook for the language’s continued survival appears bleak\cite{Norman2003}. The language faces a severe risk of extinction due to limited transmission across generations and the increasing dominance of Mandarin Chinese in daily life and formal education.

The historical development of Manchu is closely tied to the Manchu people’s rise to power and the subsequent cultural assimilation policies of the Qing Dynasty. Originally a language with no written form, Manchu adopted a script based on Mongolian in the 17th century. This script is vertically oriented and bears no resemblance to Chinese writing systems, making it unique among the major languages of the region. The script consists of syllables represented by various glyphs, creating a logographic system distinct from alphabetic languages like English.

Linguistically, Manchu is an agglutinative language with a Subject-Object-Verb (SOV) word order, meaning that verbs typically come at the end of a sentence. It employs vowel harmony and features complex morphological rules, which allow for a large variety of meanings to be expressed through prefixes and suffixes attached to a verb root. Manchu verbs, much like those in polysynthetic languages such as Cherokee, carry substantial grammatical information, including tense, aspect, mood, and voice. A typical Manchu verb may include multiple morphemes that together convey complex meanings, a feature common to languages in the Tungusic family.

A typical Manchu verb consists of several components:

\begin{itemize}
\item \textbf{Verb root}, which provides the main action or meaning of the verb
\item \textbf{Aspect markers}, indicating whether an action is completed, ongoing, or repetitive
\item \textbf{Mood markers}, which express the speaker's attitude toward the action (e.g., imperative, interrogative)
\item \textbf{Negative particles}, which are added to negate the action of the verb
\item \textbf{Directional suffixes}, specifying the direction or orientation of the action (e.g., toward or away from the speaker)
\end{itemize}

Manchu nouns, however, are simpler than verbs in terms of morphology. Unlike verbs, Manchu nouns do not have extensive inflection, and they are not marked for gender or number. However, Manchu makes a distinction between animate and inanimate entities, which is reflected in both the syntax and verb conjugation rules of the language. Additionally, Manchu nouns are often modified by possessive suffixes and directional markers, allowing for precise expression of spatial and relational concepts.

The Manchu language is critical for understanding the culture and governance of the Qing Dynasty. Thousands of historical documents, including military records, imperial edicts, and literary texts, were written in Manchu. Many of these documents remain untranslated, making proficiency in Manchu essential for historians studying Qing Dynasty China. Manchu also holds significant cultural value for the Manchu people, preserving traditional knowledge, oral histories, and religious practices.

The challenges facing the revitalization of Manchu are similar to those encountered by other endangered languages, including limited resources, a dwindling speaker base, and the overwhelming dominance of global languages like Chinese and English. However, the continued interest from scholars and the rise of community-driven language preservation efforts offer hope for the survival of this historically significant language.

\subsection{Techniques Leveraging LLMs for Low-Resource Machine Translation}

Fluent Cherokee speakers are few in number, and while there are multiple initiatives to teach the language to Cherokee school children, it is a race against time to preserve the language and the cultural knowledge it contains before another generation passes away.  Today, translation from Cherokee to English is a laborious process that involves transliterating the syllabary into the Latin Alphabet, producing a rough English analog, and then working with Cherokee elders to convey the cultural subtleties of the original text.  With the recent, rapid technological advances, however, LLMs may aid the process of translation in both directions, with the goal of producing passable texts that fluent speakers can then polish.

\subsubsection{Zero-shot and Few-shots Translation}
Due to a lack of training data for low-resource languages, most LLMs cannot properly translate low-resource languages. Moreover, it is hard to finetune these LLM without sufficient language resources. Thus, various research efforts focus on improving LLMs to perform translation tasks with minimal or no task-specific training, commonly referred to as zero-shot and few-shot learning.

Zero-shot translation refers to the capability of LLMs to translate between language pairs without having seen explicit examples of these translations during training. For instance, Zhang et al. proposed strategies such as random online back-translation and language-specific modeling, improving zero-shot performance in multilingual NMT by approximately 10 BLEU score \cite{improve_zero}. Gao et al. introduced Cross-lingual Consistency Regularization (CrossConST), which enhances zero-shot performance by bridging the representation gap across languages, proving effective in both high-resource and low-resource language settings \cite{improve_multi}. Although zero-shot translation can provide good results for certain languages, particularly when leveraging shared linguistic patterns from pretraining on multilingual corpora, it basically leverages well-resourced languages (often English) to facilitate translation between low-resource language pairs. Thus, the performance will degrade when applied to highly underrepresented languages. This highlights the need for additional strategies such as fine-tuning or incorporating domain-specific data.

Compared with no training examples in zero-shot translation, few-shot translation involves exposing a model to a small number of translation examples before performing the task. In this scenario, LLMs can adapt quickly to specific language pairs or domains with minimal data, offering a highly flexible solution for low-resource languages. Zhang et al. explored few-shot learning in machine translation using LLMs \cite{fewshot_qlora}. They compared prompting, few-shot learning, and fine-tuning approaches, finding that few-shot learning often outperforms zero-shot prompting when the model is given even a few translation examples. This research highlighted the flexibility and efficiency of LLMs, especially when fine-tuned using the QLoRA method, which allows the models to adapt to machine translation tasks with minimal data and memory usage. The study by Cahyawijaya et al. extensively explores in-context learning (ICL) and cross-lingual in-context learning (X-ICL), focusing on their application to low-resource languages \cite{llm_fewshot}. This work highlights the effectiveness of using in-context examples from high-resource languages to perform translation tasks in low-resource languages. By providing semantically aligned in-context examples, LLMs can bridge the language gap and enhance understanding of low-resource languages without fine-tuning. However, they also point out that label alignment can sometimes degrade performance in low-resource languages, and propose an alternative approach called query alignment, which focuses on aligning the input distribution rather than the labels. Additionally, Guo et al. proposed the Talent method, which utilizes a textbook-based learning approach to improve LLM translation performance on low-resource languages \cite{teach_prompt}. By creating structured textbooks containing vocabulary lists and syntax patterns, LLMs are guided to absorb this knowledge before the translation task, significantly improving performance in few-shot settings for low-resource languages

\subsubsection{Multilingual Pretrained Model}
LLMs such as mBART \cite{liu2020multilingual}, MarianMT \cite{junczys2018marian}, and T5 \cite{raffel2020exploring} are pretrained on multilingual corpora utilizing techniques such as masked language modeling (MLM) or denoising autoencoding. These approaches facilitate the learning of shared representations across diverse languages. By leveraging this multilingual knowledge, these models demonstrate an ability to transfer learned representations to low-resource languages, thereby enabling effective cross-lingual generation. This transfer learning capability enhances performance in translation and generation tasks, even for languages with limited available training data.

mBART is a sequence-to-sequence model specifically designed for multilingual tasks, with its primary training objective centered on denoising. In this framework, both the source and target language representations are jointly learned within a unified training paradigm. The model is pretrained by corrupting text sequences through a variety of noise functions (such as token masking or permutation) and then learning to reconstruct the original sequence. This approach allows mBART to develop a deep understanding of sentence structure and language semantics across multiple languages.
As a result of its pretraining strategy, mBART is capable of learning high-quality cross-lingual representations, which makes it particularly effective for translation tasks and other multilingual generation tasks. The shared encoder-decoder architecture allows for bidirectional transformation between languages, ensuring flexibility in generating text in one language based on input from another. This cross-lingual capability is further enhanced by mBART's ability to transfer learned knowledge to lower-resource languages, making it a robust tool for multilingual natural language processing tasks, even when large monolingual corpora are not available.

XLM-R (XLM-RoBERTa) \cite{conneau2019unsupervised} is a multilingual model based on the RoBERTa \cite{liu2019roberta} architecture, specifically extended to support over 100 languages. It is pretrained using a cross-lingual masked language modeling (MLM) task, which involves masking a portion of the input tokens and training the model to predict the missing tokens based on the context. This pretraining enables XLM-R to learn universal language representations that capture syntactic and semantic structures across a wide variety of languages. XLM-R inherits BERT'S optimizations and applies them to multilingual settings, allowing the model to handle a diverse set of languages with high proficiency. Its training corpus spans multiple languages so that it benefits from large-scale multilingual data to improve its understanding of both high-resource and low-resource languages.

MarianMT is a multilingual machine translation model designed specifically for efficient and high-quality translation between multiple languages. Unlike many other multilingual models, MarianMT is directly trained for translation tasks rather than relying solely on MLM or denoising pretraining. It is based on a transformer architecture that supports both many-to-one and many-to-many translation tasks. Therefore, even without the explicit source-target language pair specification, it can perform translations between a variety of languages. To achieves the general translation MarianMT utilize the shared vocabulary and token embeddings and fine-tuned on large parallel corpora. By leveraging parallel datasets during fine-tuning, MarianMT becomes adept at producing high-quality translations while maintaining efficient inference speeds. Additionally, MarianMT provides a flexible framework that can be adapted to different domain-specific translations, making it an ideal choice for machine translation applications across various industries. Its architecture ensures that MarianMT is capable of scaling to new languages and domains with minimal reconfiguration, supporting a wide range of multilingual tasks efficiently and effectively.

\subsubsection{Fine-tuning LLMs}
One effective approach to adapting a pretrained multilingual model while mitigating the risk of overfitting is the use of adapters \cite{houlsby2019parameter}. Adapters are compact, trainable neural network modules that are inserted between the layers of a pretrained model. These modules allow the model to be fine-tuned for new tasks or languages without requiring the retraining of the entire network, thus preserving the general knowledge captured during pretraining. Adapter fine-tuning is particularly advantageous for low-resource machine translation tasks, where large datasets are not readily available. Adapters enable the model to leverage the knowledge from high-resource languages while being fine-tuned on smaller, task-specific datasets, effectively reducing the risk of overfitting. By adjusting only a small number of parameters, adapters maintain the computational efficiency of the base model, making them suitable for scenarios with limited computational resources.

Among all types of fine-tuning methods, Low-Rank Adaptation (LoRA) fine-tunning \cite{hu2021lora} is one of the most successful strategies. LoRA is dessigned to reduce the computational overhead and memory requirements when fine-tuning large models. Instead of updating all the parameters of the LLM, LoRA introduces low-rank matrices into the model’s architecture, allowing adaptation of only a subset of parameters while keeping most of the model's original weights frozen. LoRA proposes an efficient method for fine-tuning large-scale pretrained models by leveraging matrix decomposition. Specifically, LoRA decomposes the weight matrices of the model into lower-dimensional matrices, and during the fine-tuning process, only these lower-dimensional matrices are updated. This significantly reducess the number of trainable parameters, enabling efficient adaptation of the model with minimal data and computational overhead. It also ensures that the fine-tuning process remains computationally efficient without compromising the model’s performance. By updating only the lower-dimensional matrices, the original model's capacity and knowledge acquired during pretraining, are largely preserved. As a result, LoRA allows for effective task-specific specialization while maintaining the benefits of the pretrained model's generalization abilities. This methodology is particularly valuable for low-resource machine translation (MT) tasks, where large amounts of parallel data are unavailable. LoRA’s ability to adapt models efficiently makes it ideal for such scenarios, enabling the model to specialize in low-resource language pairs without the need for full-scale retraining. 

\subsubsection{Back-Translation}
To overcome the scarce availability of large parallel corpora researchers have developed back-translation, a widely adopted data augmentation technique that leverages monolingual data to improve translation performance. Back-translation is a semi-supervised learning technique used in NMT to generate additional synthetic parallel data. The process involves training a model to translate from the target language to the source language and using this model to translate monolingual data in the target language back to the source language. This synthetic source-target data pair is then used to augment the original parallel dataset, thus improving the performance of the translation model in the source-to-target direction. LLMs can also leverage back-translation by fine-tuning a model to first generate high-resource language translations. Since evaluating the accuracy and quality of high-resource languages is significantly more feasible compared to low-resource languages, this approach ensures the generation of abundant, high-quality augmented data for low-resource language tasks. Among Back-Translation, several variations of back-translation have been developed to enhance its effectiveness. 

The most common and straightforward form of back-translation involves using a single translation model to generate synthetic source sentences from monolingual target data. Previously this model is simple but suffers from overfitting to the synthetic data if not properly balanced with authentic parallel data. However, LLMs serve as a highly effective alternative due to their extensive and versatile knowledge base, which enables them to outperform previous simple model with high accuracy and adaptability.

Iterative back-translation is an extension of the basic technique where the forward and backward models are trained in tandem, improving each other’s performance over successive iterations. It involves several steps:(1) Train an initial forward translation model from source to target language using the available parallel data. (2) Train the reverse model from target to source using the parallel data. (3) Generate synthetic data using the reverse model and retrain the forward model with the augmented dataset. (4) Generate synthetic target sentences using the updated forward model and retrain the reverse model with this new data. This iterative process continues until the model performance converges, leading to better generalization as both models reinforce each other through the generation of progressively higher-quality synthetic data. Fine-tuned LLMs are particularly well-suited for iterative back-translation due to their ability to capture complex linguistic patterns and context across languages. Their extensive pre-trained knowledge allows for nuanced understanding and generation of text, making them ideal for producing accurate and contextually appropriate translations, even in low-resource language settings. Additionally, their adaptability enables continuous improvement through fine-tuning, further enhancing their performance in iterative back-translation tasks.

\subsubsection{Retrieval-Augmented Generation}
Retrieval-Augmented Generation (RAG) typically involves three key stages: Indexing, Retrieval, and Generation \citep{lewis2020retrieval,ma2023query}.
Indexing aims to transform raw data into a vectorized database. This stage begins by converting external data—ranging from structured to unstructured formats—into a standardized format. An embedding model then encodes the processed data into smaller chunks, which are stored in the vectorized database. Indexing establishes a robust foundation for efficient and precise retrieval.
Retrieval involves applying RAG algorithms to search for and expand the prompt. In this stage, the user's initial query is encoded into an input vector using the same embedding model utilized during Indexing. The system then computes the similarity between the input vector and the stored chunks, selecting the top K most relevant chunks based on their proximity.
Generation utilizes a synthetic prompt formed by combining the user’s query with the retrieved documents. This enriched prompt provides the LLM with additional contextual information, which helps mitigate hallucinations or constrain the generated responses. Consequently, the Generation stage enhances the accuracy and reliability of the model's outputs.

Advanced RAG methods have been developed to address the limitations of naive RAG by optimizing various components, including query prompts \citep{ma2023query,peng2024large,gao2022precise}, indexing structures, similarity calculations, and prompt integration. Additionally, some advanced approaches leverage retrieval data to enhance training datasets, particularly in low-resource domains.

For instance, \citep{seo2024retrieval} retrieves relevant instances and uses large language models (LLMs) to generate new samples that integrate both original and retrieved data, thereby addressing data scarcity in specialized domains. \citep{parvez2022retrieval} involves utilizing retriever models to identify relevant samples and expand the set of positive examples within privacy policy question-answering datasets, enhancing the training process. However, these methods often rely on external data, which poses challenges when the data is sensitive and cannot be accessed due to privacy concerns, or when computational resources are limited.

\subsection{LLMs Empowered Machine Translation Task}
\subsubsection{Methodology}
\begin{figure}[H]
  \centering
  \includegraphics[width=1.0\linewidth]{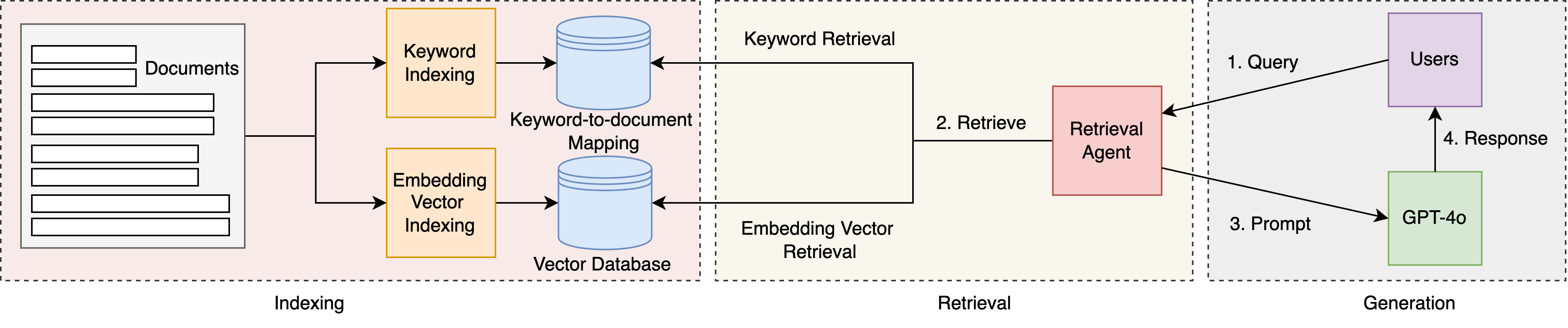}
  \caption{Illustrating a retrieval-augmented generation (RAG) architecture: Documents are indexed using both keyword and embedding vector methods, stored in separate databases. A retrieval agent accesses these indexes to provide relevant information, which is then processed by a GPT-4 model to deliver responses to users.}
  \label{fig:rag}
\end{figure}

We implement a RAG LLM low resource language translator by combining retrieval-based techniques with LLMs to ensure accurate and context-aware translations. The overall architecture is shown in Figure \ref{fig:rag}. The system utilizes dictionary entries, which are indexed through two complementary approaches: keyword-to-document mappings and vector embeddings. Key-to-document mappings in systems refer to a process where keywords are linked directly to the documents or data entries that contain or are relevant to those keywords. The keyword retriever will retrieve corresponding documents according to the key-to-document mapping, if the keyword is inside our storage. The vector embedding indexing process organizes raw linguistic data into retrievable units by associating words with dictionary definitions and using text-embedding-ada-002 model to encode the text into high-dimensional vectors that capture semantic relationships beyond mere surface forms.
Our model operates with this dual retrieval mechanism. First, a keyword-based index allows for fast and efficient lookup by identifying exact matches between query terms and dictionary entries. This method ensures that direct translations of words or phrases are retrieved whenever possible. Second, in cases where no exact keyword matches are found, the system employs a vector-based retrieval method using cosine similarity. This approach encodes the query into a semantic vector and calculates similarity scores between the query vector and all indexed vectors. The top K most similar entries are then retrieved, ensuring that semantically related content such as synonyms or conceptually similar expressions are identified, even in the absence of exact matches.

After retrieval, the system uses a GPT-based model to synthesize the retrieved content into a coherent and fluent translation. This generative step integrates the context from retrieved entries to resolve ambiguities and produce a natural, human-readable translation. By balancing lightweight keyword-based retrieval with deeper semantic understanding through vector-based retrieval, this hybrid approach enhances both the accuracy and relevance of the translation output. The RAG architecture thus combines precise keyword matching with the flexibility of semantic search, making it particularly effective for low-resource languages like Cherokee.

\subsubsection{Testing Data}
Given the extreme scarcity of parallel translations between Cherokee and English literary works, this experiment utilizes three Cherokee literary sources that have available translations: the New Testament, Peter Parley's Geography, and The Pilgrim's Progress. These texts provide a rare opportunity to assess translation performance, as they offer both Cherokee and English versions for comparative analysis.

Cherokee New Testament is the largest single book written in Cherokee. This translation of the New Testament of the Bible into the Cherokee language, using the Cherokee syllabary, was completed in the early 19th century. It contains the verse written in Cherokee using Cherokee syllabary and Cherokee Latin transliteration, which provides the pronunciation of the Cherokee syllabary text using Latin letters. The translation was spearheaded by Samuel Worcester, a missionary, and Elias Boudinot, a prominent Cherokee, with significant involvement from Cherokee speakers. The Cherokee New Testament was printed and widely distributed by the Cherokee Nation and became a vital text in preserving the Cherokee language and literacy among Cherokee people. Given its importance in Cherokee history and its length compared to other works in Cherokee, it is considered one of the most substantial works written in the language.

Peter Parley's Geography, written by Samuel Griswold Goodrich under the pseudonym Peter Parley, was a popular 19th-century educational book designed to teach geography to young readers. First published in the early 1820s, the book introduces geographical concepts in an accessible and engaging manner, using illustrations, maps, and simple language to explain the physical world, cultures, and places beyond the reader's immediate surroundings. Notably, Peter Parley's Geography was one of the earliest geography textbooks aimed at children, emphasizing both geographical knowledge and moral lessons. Its influence was significant during the 19th century, and its translation into Cherokee exemplifies the efforts to provide educational material to indigenous populations, reflecting a period of educational reform and expansion.

The Pilgrim's Progress is one of the most significant works in English literature and Christian allegory written by John Bunyan and first published in 1678. The book narrates the journey of its protagonist, Christian, from his home in the "City of Destruction" to the "Celestial City," symbolizing the believer's path to salvation. Divided into two parts, the story combines spiritual lessons with vivid characters and settings, reflecting the struggles, temptations, and rewards of living a faithful Christian life.

We conducted the translation for a transcript from the Harris-Trump presidential debate to assess the performance of our model in the context of news and current events \cite{president}. This evaluation aimed to determine how well the system could handle complex, real-world content and provide timely information to Cherokee speakers. The debate transcript was chosen for its rich linguistic features, including political terminology, colloquial expressions, and nuanced argumentation.

We also utilize the New Testament to evaluate our model on Tibetan, a language of significant historical and cultural importance in Asia, and Manchu, a critically endangered language with less than 100 speakers\cite{Weers2016}. Due to a lack of resources, the Bible is one of the only available materials across Cherokee, Manchu, and Tibetan languages. We test Tibetan with This evaluation seeks to determine whether our model is capable of effectively handling other low-resource languages, further validating its generalizability and performance beyond Cherokee. By extending our analysis to these languages, we aim to explore the model’s robustness in addressing the unique challenges posed by different low-resource linguistic contexts.
\subsection{Test Design and Evaluation}

To evaluate the performance and capabilities of our model and original LLMs in low-resource machine translation, we conducted an experiment comparing our model with GPT-4o and LLaMA 3.1 405B models. The objective of the experiment was to assess the translation abilities of these models in generating low-resource languages, focusing primarily on Cherokee translations from English source sentences. Specifically, we randomly selected text from three test datasets for Cherokee and New Testament for Tibetan and Manchu with comparable sentence lengths. Each sentence was input sequentially into each model, which was tasked with translating the sentences into low-resource languages. A comprehensive evaluation framework was then applied to assess and analyze the performance of these models, allowing for a detailed comparison and discussion of their effectiveness in low-resource language generation. We employed a combination of fundamental evaluation metrics, including ROUGE and BLEU, alongside advanced semantic evaluation techniques such as BERTScore and human expert assessment. This multi-faceted evaluation approach allows for both quantitative and qualitative analysis of the generated translations, providing a comprehensive assessment of the models' performance in terms of both lexical accuracy and semantic fidelity.

ROUGE (Recall-Oriented Understudy for Gisting Evaluation) is primarily used to evaluate the similarity between the machine-generated translations and reference translations by measuring overlapping n-grams, word sequences, and word pairs. It is commonly used in summarization tasks but also provides insights into the fluency and accuracy of translations.

BLEU (Bilingual Evaluation Understudy) is another widely adopted metric in machine translation, which calculates the precision of n-grams between the generated translations and reference translations. It provides a score that reflects how many words or phrases in the generated output are found in the reference, making it a valuable metric for assessing the fidelity of translation.

In addition to these surface-level metrics, we applied BERTScore, which leverages contextual embeddings from pretrained models to capture semantic similarity between the generated and reference translations. Unlike ROUGE and BLEU, which focus on exact n-gram matches, BERTScore evaluates the meaning and context, making it particularly useful for low-resource language tasks where exact matches may not fully capture translation quality.

In addition to automated metrics, we incorporated human expert evaluation to provide a qualitative assessment of the generated translations. The expert evaluated the translations based on three key dimensions: fluency, grammaticality, and faithfulness to the source text. Each of these aspects is scored on a scale from 0 to 5, with higher scores indicating better performance. Fluency measures the naturalness and ease of reading in the target language; grammaticality assesses adherence to the syntactic rules of Cherokee. Faithfulness evaluates how accurately the translation conveys the meaning of the original English sentence. To ensure consistency with other scoring metrics, we will standardize the expert evaluation scores to a range of 0 to 1. This will be achieved by summing all the scores across fluency, grammaticality, and faithfulness, then dividing by the maximum possible total score. This approach provides a consistent scale, allowing for more straightforward comparisons across translations. The formula for the normalized human evaluation score is shown as following:
\begin{equation}
\text{Human Evaluation} = \frac{1}{n}\sum_{i=1}^{n} \frac{ \text{Fluency}_i + \text{Grammaticality}_i + \text{Faithfulness}_i}{5 \times 3}
\end{equation}
In this formula, $n$ is the number of sentences, and the sum of full grade of fluency, grammaticality and faithfulness, 5 point each, is $15$ in total.
Fluency$_i$, Grammaticality$_i$, and Faithfulness$_i$ are the expert scores for sentence $i$ in each respective dimension.

\section{Experiment Results}

\begin{figure}[H]
  \centering
  \includegraphics[trim=100 140 120 110,clip, width=.8\linewidth]{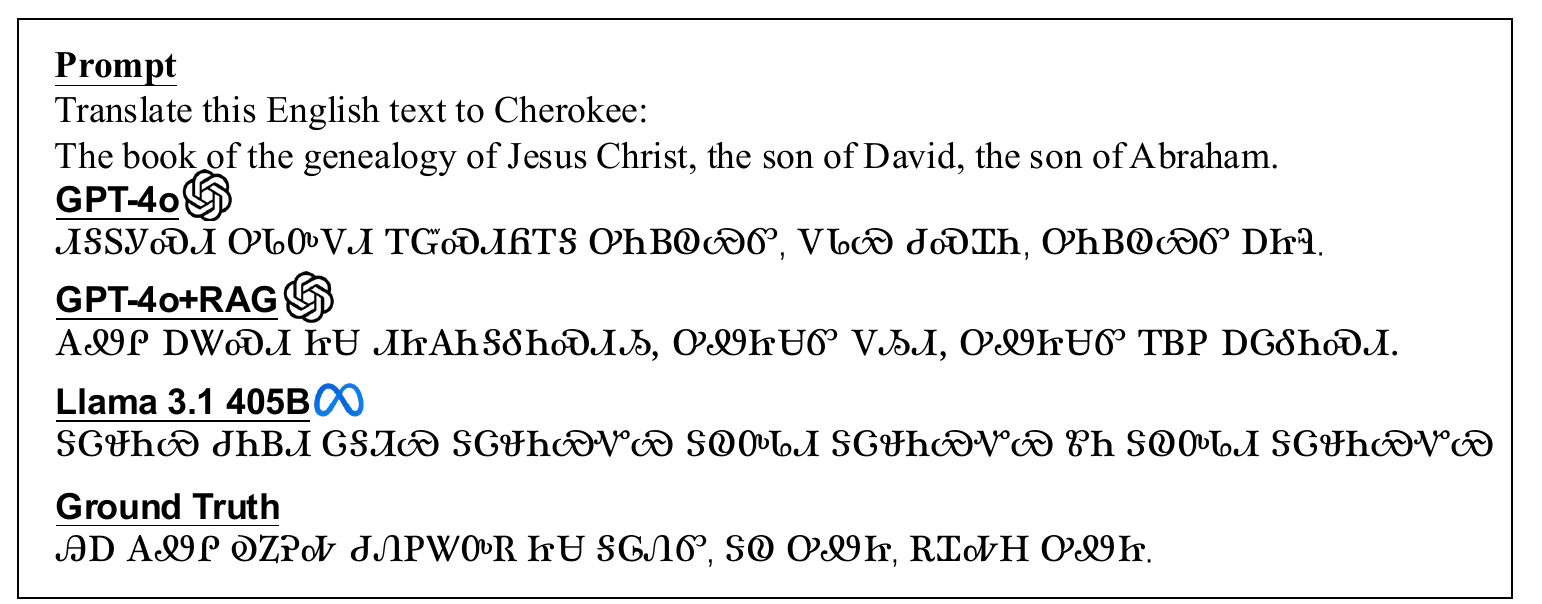}
  \caption{One test example on Cherokee New Testament given the ground truth translation. We test GPT-4o, GPT-4o with RAG and Llama 405B.}
  \label{fig:1}
\end{figure}

\begin{figure}[H]
  \centering
  \includegraphics[trim=50 30 10 13,clip,width=1.0\linewidth]{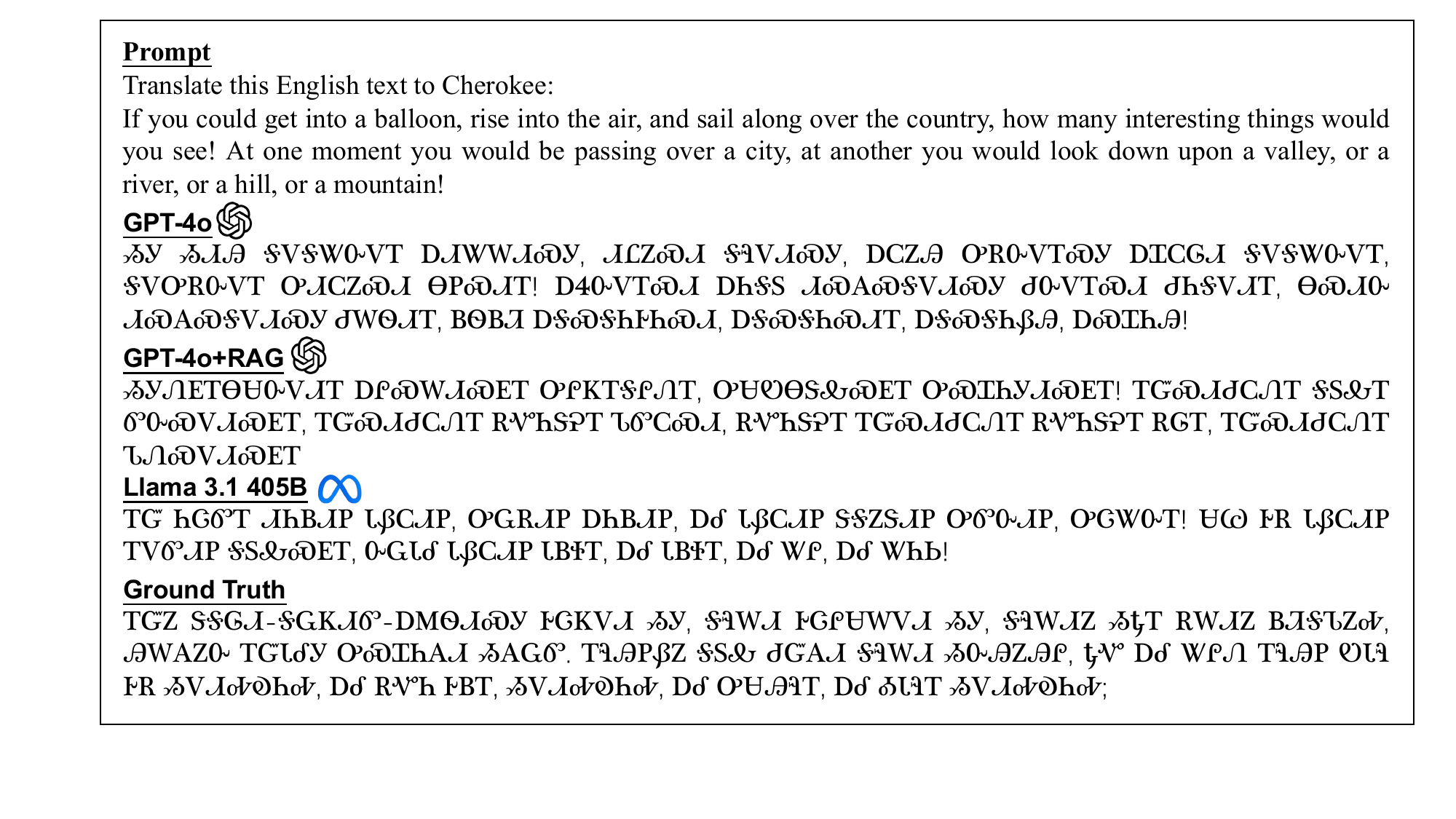}
  \caption{One example for testing LLMs translation on Peter Parley’s Geography.}
  \label{fig:11}
\end{figure}

\begin{figure}[H]
  \centering
  \includegraphics[trim=15 20 30 15,clip,width=1.0\linewidth]{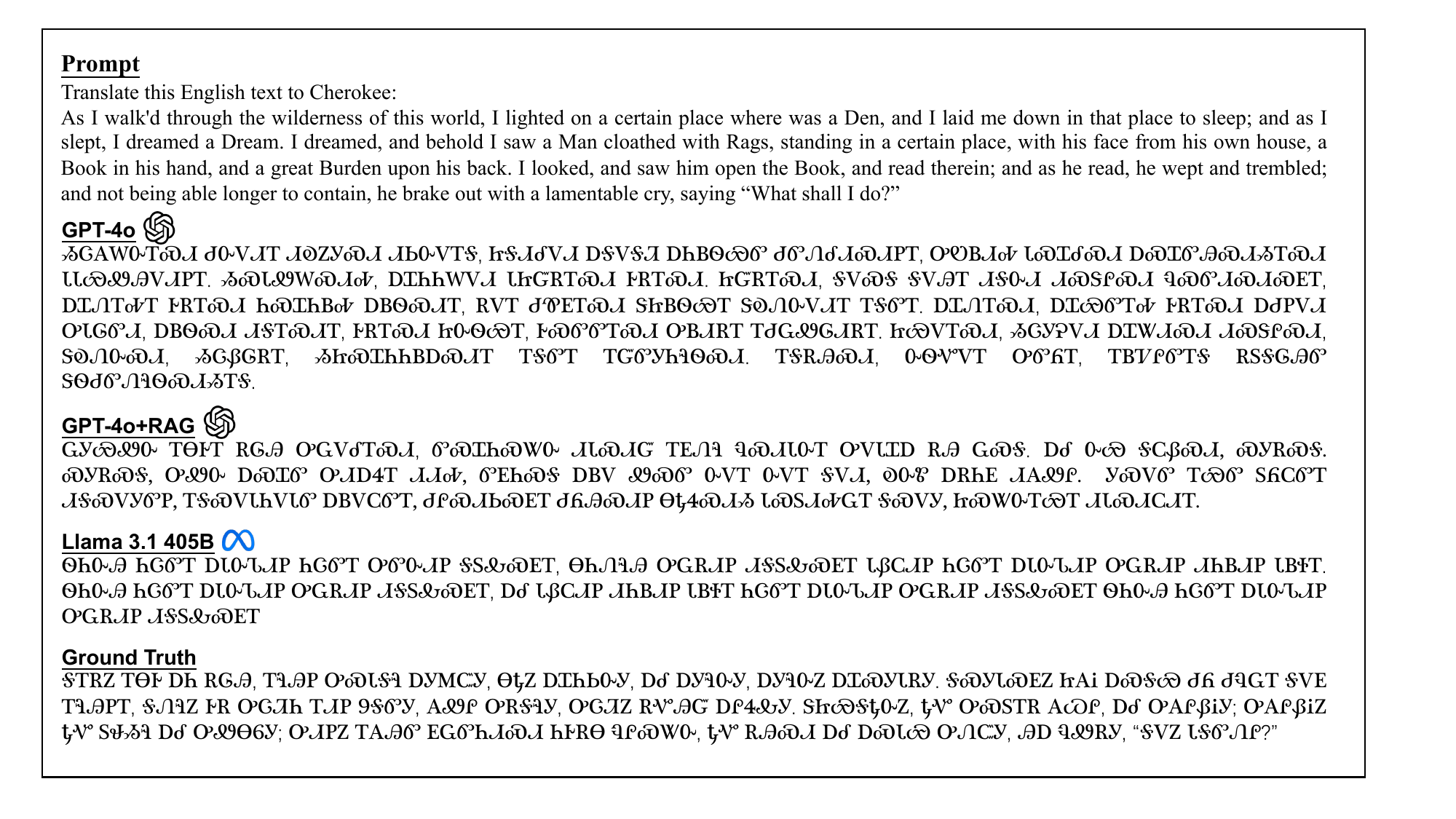}
  \caption{One example for testing LLMs translation on The Pilgrim’s Progress.}
  \label{fig:16}
\end{figure}

\begin{figure}[H]
  \centering
  \includegraphics[trim=15 20 15 24,clip,width=1.0\linewidth]{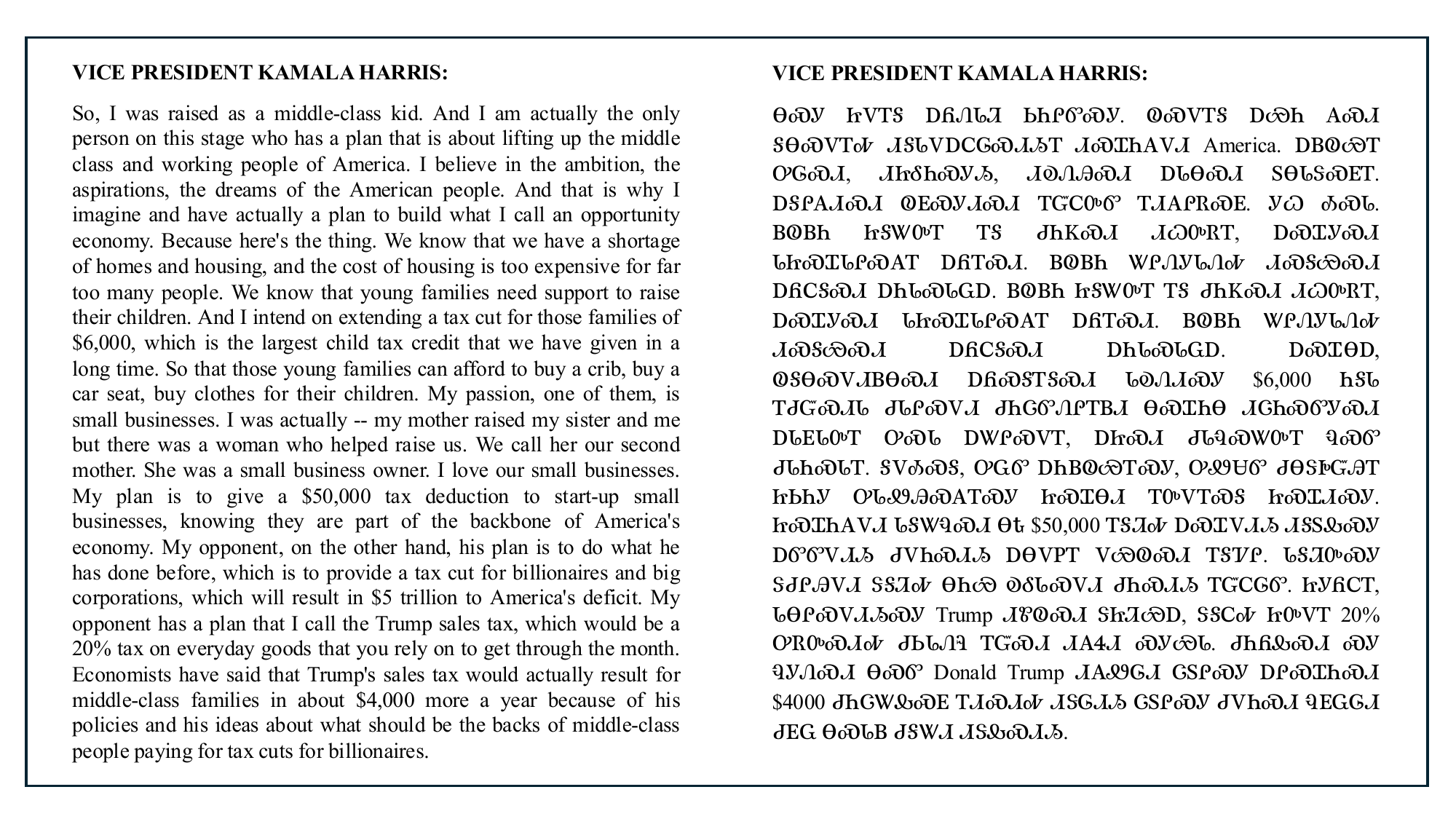}
  \caption{We applied our retrieval-based model to translate a paragraph from the Kamala Harris debate, presenting the original English text on the left and the corresponding Cherokee translation on the right.}
  \label{fig:Harris}
\end{figure}

\begin{figure}[H]
  \centering
  \includegraphics[width=1.0\linewidth]{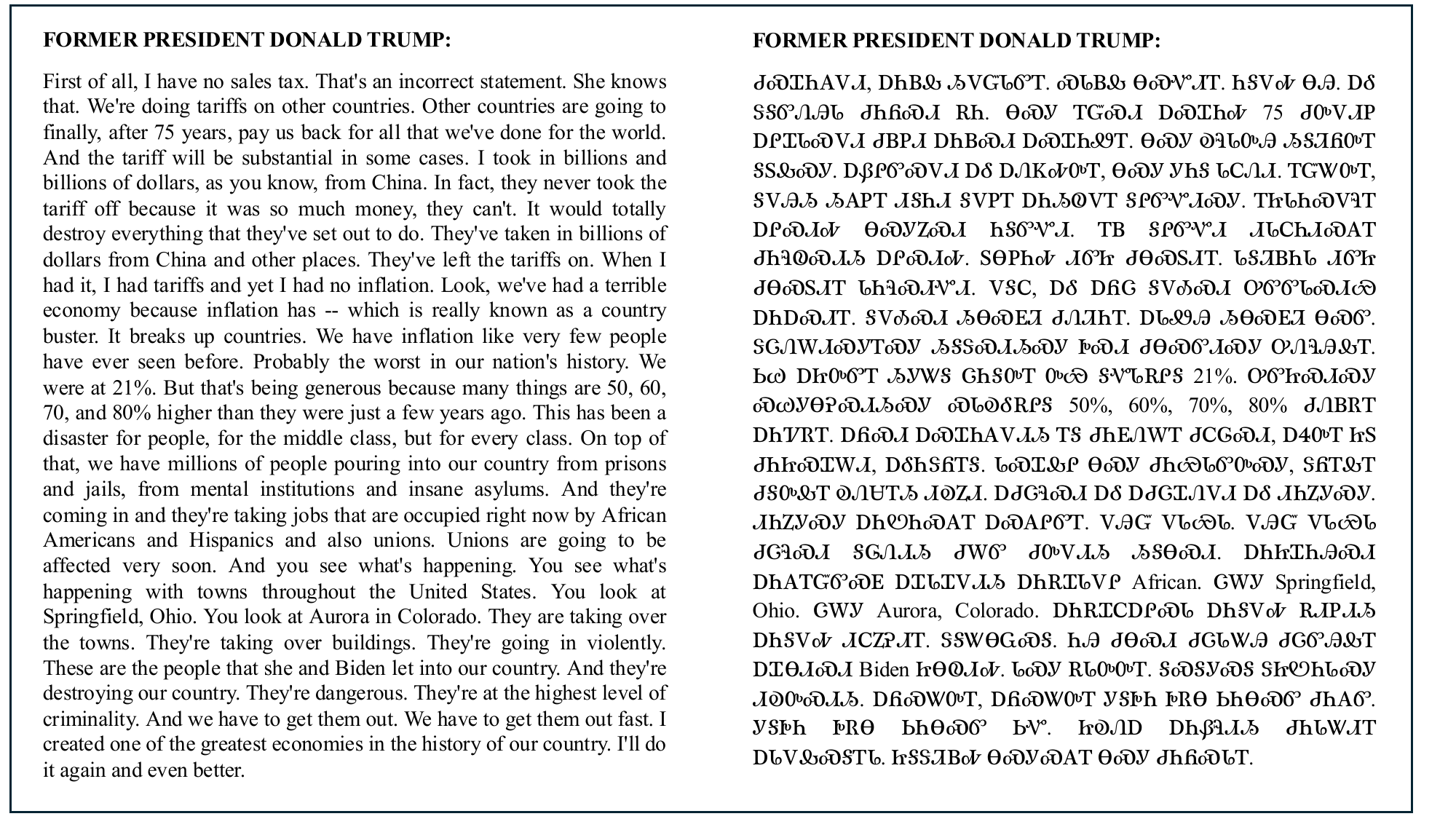}
  \caption{We applied our retrieval-based model to translate a paragraph from the Donald Trump debate, presenting the original English text on the left and the corresponding Cherokee translation on the right.}
  \label{fig:Trump}
\end{figure}

\begin{figure}[H]
  \centering
  \includegraphics[trim=160 90 90 24,clip,width=\linewidth]{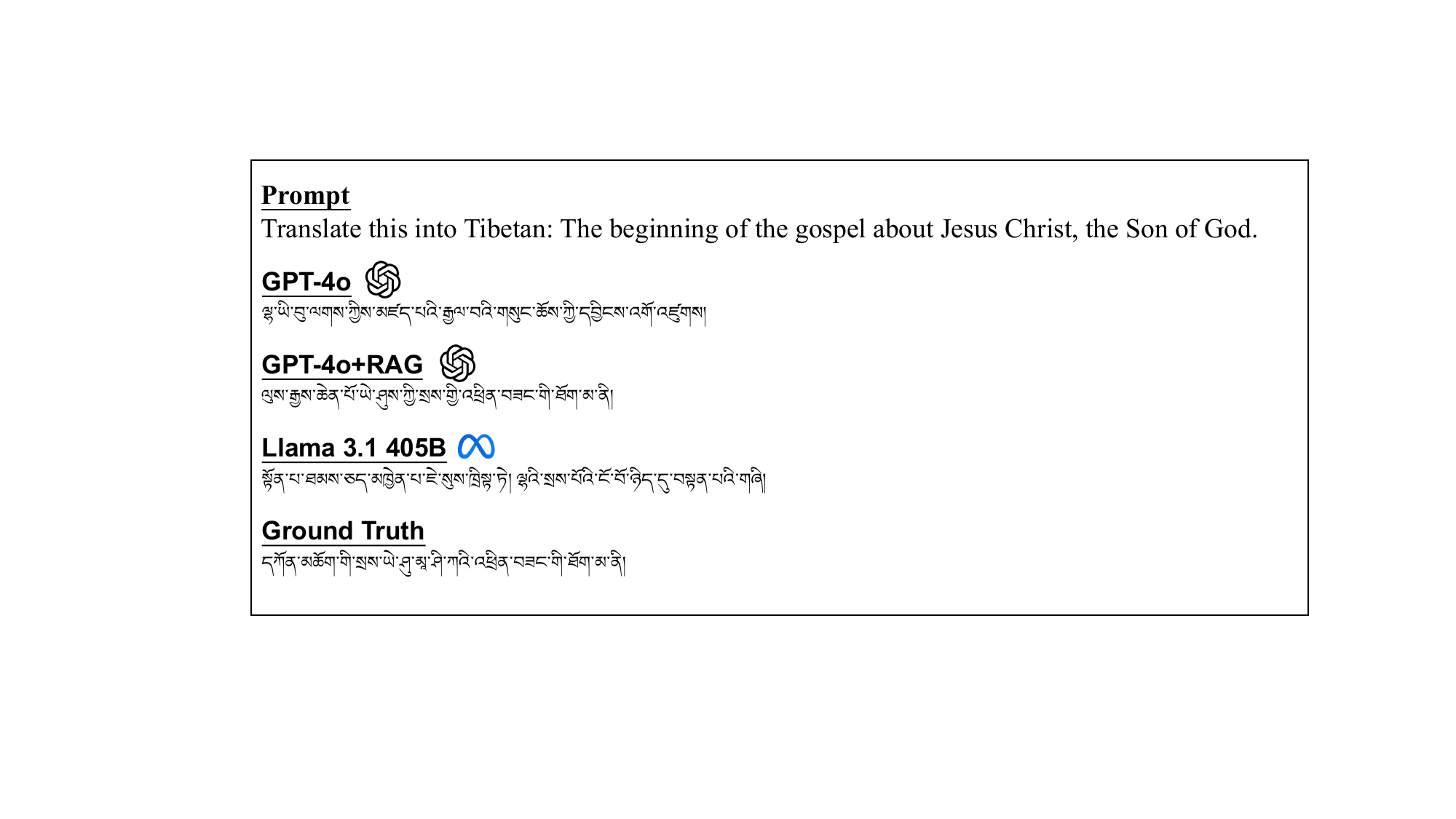}
  \caption{One example for testing LLMs translation for Tibetan.}
  \label{fig:21}
\end{figure}

\begin{figure}[H]
  \centering
  \includegraphics[trim=200 30 160 24,clip,width=\linewidth]{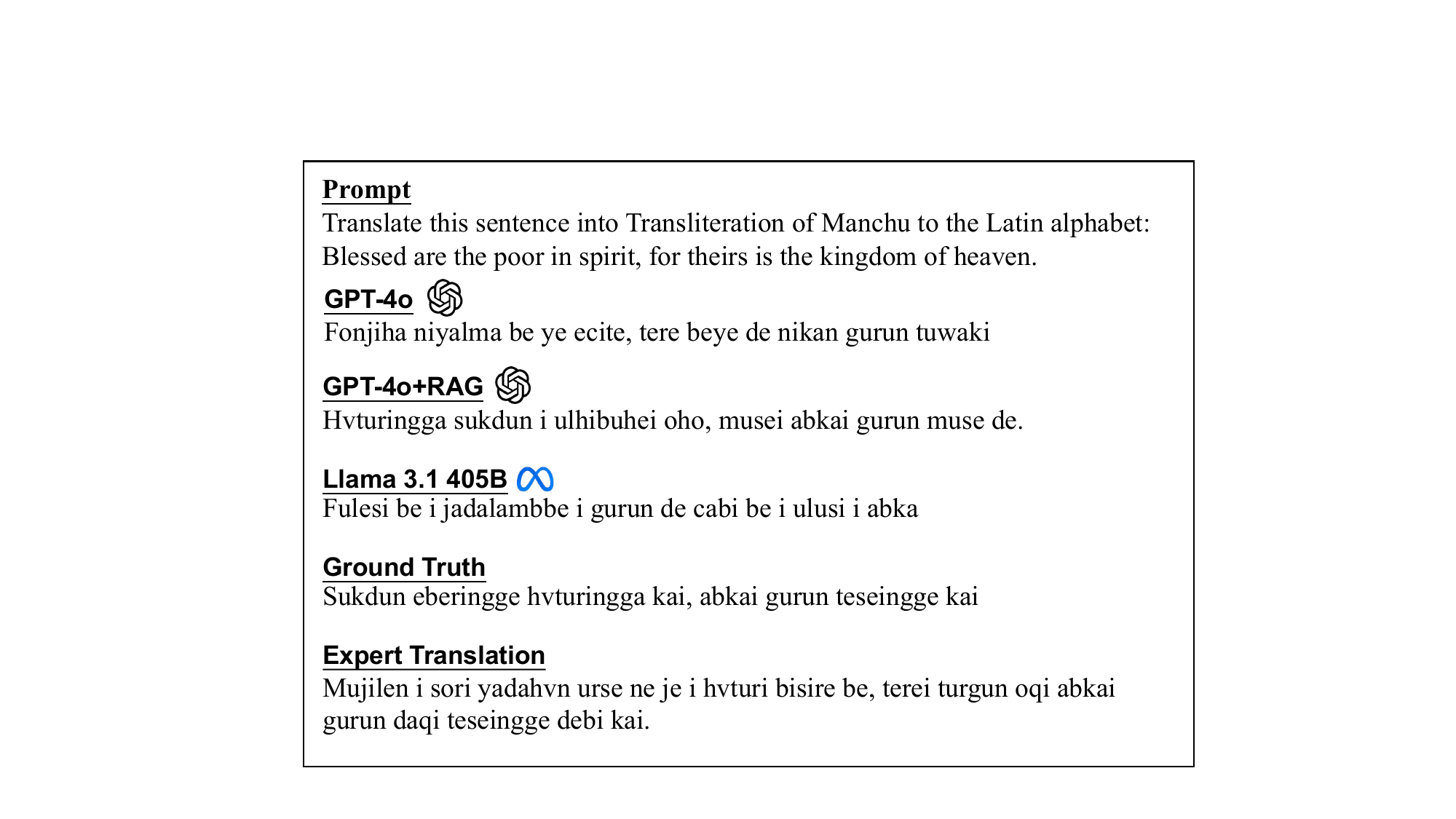}
  \caption{One example for testing LLMs translation for Manchu.}
  \label{fig:26}
\end{figure}

We show our evaluation results in table \ref{tab:experiment_result}. Figure \ref{fig:1} to \ref{fig:16} show three examples applying LLMs for each Cherokee literary source. We then exhibit translation of 2024 president debate for both presidential candidates in figure \ref{fig:Harris} and \ref{fig:Trump}. Finally, figure \ref{fig:21} and \ref{fig:26} include one test example for Tibetan and Manchu translation. Following \cite{seo2023mergen}, we use the transliteration of Manchu to the Latin alphabet when generating the Manchu translation. The complete evaluation examples and more details are shown in the appendix \ref{additional}.

\begin{table}[]
\centering
\resizebox{\columnwidth}{!}{%
\begin{tabular}{@{}cccccccc@{}}
\toprule
Language & Model & BLEU & ROUGE-L & \begin{tabular}[c]{@{}c@{}}BERTScore\\ Precision\end{tabular} & \begin{tabular}[c]{@{}c@{}}BERTScore\\ Recall\end{tabular} & \begin{tabular}[c]{@{}c@{}}BERTScore\\ F1\end{tabular} & Human Evaluation \\ \midrule
& Llama 3.1 405B & 0.0 & 0.0 & 0.931 & 0.927 & 0.929 & 0.0 \\
Cherokee & GPT-4o & 0.003 & 0.0 & 0.938 & 0.938 & 0.938 & 0.0 \\
& GPT-4o + RAG & 0.115 & 0.117 & 0.962 & 0.964 & 0.963 & 0.0 \\
\midrule
& Llama 3.1 405B & 0.0 & 0.0 & 0.879 & 0.859 & 0.869 & 0.067 \\
Tibetan & GPT-4o & 0.0 & 0.0 & 0.833 & 0.851 & 0.842 & 0.147 \\
& GPT-4o + RAG & 0.108 & 0.123 & 0.802 & 0.810 & 0.806 & 0.293 \\
\midrule
& Llama 3.1 405B & 0.0 & 0.104 & 0.693 & 0.663 & 0.678 & 0.040 \\
Manchu & GPT-4o & 0.0 & 0.125 & 0.726 & 0.703 & 0.714 & 0.173 \\
& GPT-4o + RAG & 0.077 & 0.188 & 0.716 & 0.696 & 0.706 & 0.333 \\
\bottomrule
\end{tabular}%
}
\caption{Experimental results comparing our model with GPT-4o, GPT-4o with RAG and LLaMA 3.1 405B across three different low-resource languages. We only test New Testament translation in Tibetan and Manchu experiments.}
\label{tab:experiment_result}
\end{table}

\begin{table}[]
\centering
\resizebox{\columnwidth}{!}{%
\begin{tabular}{@{}cc|ccc|ccc|ccc@{}}
\toprule
Model & & \begin{tabular}[c]{@{}c@{}}\end{tabular} & \begin{tabular}[c]{@{}c@{}}Llama 3.1 405B\end{tabular} & \begin{tabular}[c]{@{}c@{}}\end{tabular} & \begin{tabular}[c]{@{}c@{}}\end{tabular} & \begin{tabular}[c]{@{}c@{}}GPT-4o\end{tabular} & \begin{tabular}[c]{@{}c@{}}\end{tabular} & \begin{tabular}[c]{@{}c@{}}\end{tabular} & \begin{tabular}[c]{@{}c@{}}GPT-4o +RAG\end{tabular} & \begin{tabular}[c]{@{}c@{}}\end{tabular} \\
\midrule
Languages & Sentence Index & \begin{tabular}[c]{@{}c@{}} Fluency\end{tabular} & \begin{tabular}[c]{@{}c@{}}Grammaticality\end{tabular} & \begin{tabular}[c]{@{}c@{}}Faithfulness\end{tabular} & \begin{tabular}[c]{@{}c@{}}Fluency\end{tabular} & \begin{tabular}[c]{@{}c@{}}Grammaticality\end{tabular} & \begin{tabular}[c]{@{}c@{}} Faithfulness\end{tabular} & \begin{tabular}[c]{@{}c@{}} Fluency\end{tabular} & \begin{tabular}[c]{@{}c@{}}Grammaticality\end{tabular} & \begin{tabular}[c]{@{}c@{}}Faithfulness\end{tabular} \\ 
\midrule
 & \ref{fig:1} & 0 & 0 & 0 & 0 & 0 & 0 & 0 & 0 & 0 \\
 & \ref{fig:2} & 0 & 0 & 0 & 0 & 0 & 0 & 0 & 0 & 0 \\
 & \ref{fig:3} & 0 & 0 & 0 & 0 & 0 & 0 & 0 & 0 & 0 \\
 & \ref{fig:4} & 0 & 0 & 0 & 0 & 0 & 0 & 0 & 0 & 0 \\
\multirow{-5}{*}{Cherokee} & \ref{fig:5} & 0 & 0 & 0 & 0 & 0 & 0 & 0 & 0 & 0 \\ \midrule
& \ref{fig:21} & 1 & 0 & 1 & 1 & 1 & 1 & 1 & 1 & 2 \\
 & \ref{fig:22} & 0 & 1 & 0 & 0 & 0 & 1 & 1 & 1 & 2 \\
 & \ref{fig:23} & 0 & 0 & 0 & 1 & 0 & 1 & 0 & 1 & 1 \\
 & \ref{fig:24} & 0 & 0 & 1 & 0 & 0 & 1 & 1 & 2 & 2 \\
\multirow{-5}{*}{Tibetan} & \ref{fig:25} & 0 & 0 & 1 & 1 & 2 & 1 & 2 & 2 & 3 \\ \midrule
 & \ref{fig:26} & { 0} & { 0} & { 1} & { 0} & { 1} & { 1} & { 1} & { 1} & { 1} \\
 & \ref{fig:27} & { 0} & { 0} & { 1} & { 0} & { 0} & { 1} & { 1} & { 2} & { 2} \\
 & \ref{fig:28} & { 0} & { 0} & { 0} & { 1} & { 0} & { 1} & { 2} & { 2} & { 3} \\
 & \ref{fig:29} & { 0} & { 0} & { 0} & { 2} & { 1} & { 1} & { 3} & { 2} & { 2} \\
\multirow{-5}{*}{Manchu} & \ref{fig:30} & { 0} & { 1} & { 0} & { 2} & { 1} & { 1} & { 3} & { 2} & { 2} \\ \bottomrule
\end{tabular}%
}
\caption{Human experts evaluation scores for Llama 3.1 405B, GPT-4o and GPT-4o + RAG on three low resources languages. The sentence index refers to the examples shown in the paper.}
\label{tab:human_evaluation}
\end{table}

Both LLaMA 3.1 405B and GPT-4o exhibit notably low scores in traditional machine translation evaluation metrics such as BLEU and ROUGE-L. These results indicate a near-complete lack of overlap between the generated translations and the reference texts in terms of both word n-gram precision measured by BLEU and sentence-level structure measured by ROUGE-L. Such results suggest that both models struggle significantly with the low-resource language translation task when evaluated using metrics that prioritize surface-level textual similarity. The zero scores in ROUGE-L across both models highlight their inability to produce outputs that match the structural or syntactic form of the reference translations, due to the inherent challenges associated with the limited availability of Cherokee training data and the unique linguistic structure of Cherokee compared to high-resource languages.

However, the GPT-4o+RAG model demonstrates a significant improvement over the standalone models in terms of BLEU and ROUGE-L scores. This suggests that the integration of RAG provides the model with additional contextual information that helps align its output more closely with the reference translations in terms of both lexical overlap and syntactic structure. The higher scores indicate that the retrieval component allows the model to generate translations that have better word and phrase alignment with the reference texts, addressing some of the structural limitations seen in the other models.

Despite the poor performance in BLEU and ROUGE-L for LLaMA 3.1 405B and GPT-4o, all models demonstrate significantly higher scores when evaluated using BERTScore, which focuses on semantic similarity rather than exact word matches. GPT-4o+RAG outperforms both LLaMA 3.1 405B and GPT-4o across all BERTScore metrics. These high values indicate that the GPT-4o+RAG model is the most effective in retaining semantic meaning in its translations. Its retrieval mechanism not only enhances the model's lexical alignment but also strengthens its ability to preserve the semantic essence of the source text.

The human evaluation results reveal the models' ability to produce translations that align with human expectations of quality. For LLaMA 3.1 405B and GPT-4o models, human evaluation scores remain relative low across all three languages. Likewise, the GPT-4o + RAG model achieves the highest human evaluation scores in both Tibetan and Manchu, indicating a notable improvement in aligning translations with human expectations compared to the standalone models. However, all three models receive a score of 0 in Cherokee translation, even for GPT-4o + RAG model, which otherwise performs best on BLEU, ROUGE-L, and BERTScore metrics. This result suggests that the model’s output may still lack cultural and contextual fidelity essential in low-resource language translation of Cherokee Language. 

The low human evaluation scores for Cherokee translations can be partly explained by the relatively recent development of a written system for Cherokee compared to the long textual histories of Manchu and Tibetan. Cherokee’s writing system was only created in the early 19th century by Sequoyah\cite{cushman2011}, making it a young written language with fewer established conventions and limited historical literature. This lack of reference material restricts the models’ ability to learn from and align with standard expressions and syntactic norms. These findings indicate that while RAG improves the surface-level and semantic alignment, further advancements are needed to capture the nuanced syntactic and cultural elements that human evaluators prioritize in translation quality, emphasizing the value of integrating culturally-aware contextual knowledge into the translation models.

In summary, while the LLaMA 3.1 405B and GPT-4o models show considerable limitations in their ability to translate Cherokee with lexical and structural accuracy, the GPT-4o+RAG model provides a more promising approach by leveraging external information to improve translation quality. The significant disparity between BLEU/ROUGE-L and BERTScore for all models suggests that these models, even with RAG integration, still face challenges in achieving lexical and syntactic fidelity to the reference translations. Nonetheless, the relatively high BERTScore values for GPT-4o+RAG imply that our approach is better suited for low-resource translation tasks, offering a more balanced solution that captures both structural and semantic nuances more effectively.

\section{Discussions \& Conclusion}

Preserving minority languages, such as Cherokee, is important for maintaining cultural heritage and linguistic diversity. However, in the digital age, these languages often face challenges due to a lack of resources and training data, which exacerbates the risk of their decline. To address this, AI models that generate high-quality translations could play a crucial role in sustaining minority languages \cite{wang2023seeing,wang2024system}. This digital integration would not only support language preservation but also could revitalize interest among younger generations, ensuring these unique linguistic perspectives continue to enrich our global discourse.

In this study, we evaluated the performance of our retrieval-based model alongside GPT-4o and Llama 3.1 405B in translating English into Cherokee, Tibetan, and Manchu. Our findings demonstrated a significant improvement in translation accuracy with our method compared to GPT-4o and Llama 3.1 405B, underscoring the effectiveness of our approach. AI models can effectively translate low-resource languages, they could assist in revitalization efforts by providing learners with more reliable language materials and tools for practice. Furthermore, integrating AI-driven translation into language learning programs could help bridge gaps in resources, supporting both native speakers and new learners. This could lead to more sustainable language preservation, helping minority languages remain active in both spoken and written forms. Overall, this study demonstrates the potential of AI to bridge the gap between traditional knowledge and modern technology in the preservation of minority languages. We hope this study can inspire further research and innovation in language revitalization efforts, and help minority languages continue to thrive, preserving unique worldviews and cultural perspectives for future generations.

\bibliographystyle{splncs04}
\bibliography{mybib} 

\newpage

\appendix
\section{Appendix}
\subsection{Additional Experiment Results}
\label{additional}
In this section, we present additional experimental results based on translations of both short and long texts to further evaluate the performance of our model, GPT-4o and LLaMA 3.1 405B in low-resource language translation. Specifically, we tasked the models with translating Chapter 1 of the Gospel of Matthew from the New Testament sentence by sentence to assess their performance on concise, context-rich language. For Peter Parley’s Geography, we chose a single paragraph that provides geographical descriptions, while for The Pilgrim’s Progress, we similarly selected a paragraph. We gave relatively long sentence for both to evaluate how the models handle longer and more continuous narrative found in texts.

\begin{figure}[H]
  \centering
  \includegraphics[trim=30 80 30 80,clip,width=\linewidth]{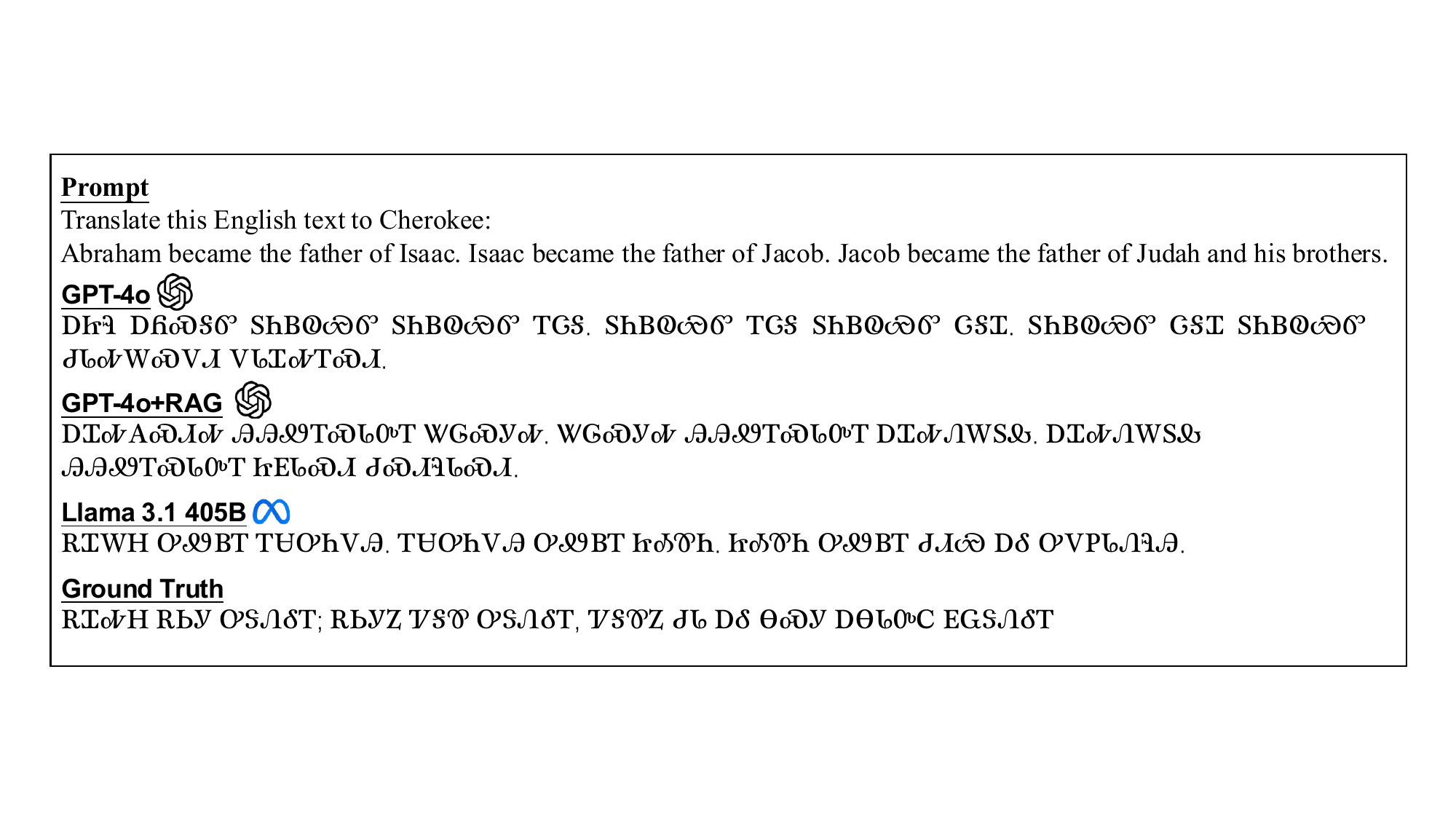}
  \caption{Additional testing example on Cherokee New Testament.}
  \label{fig:2}
\end{figure}

\begin{figure}[H]
  \centering
  \includegraphics[trim=20 100 10 80,clip,width=\linewidth]{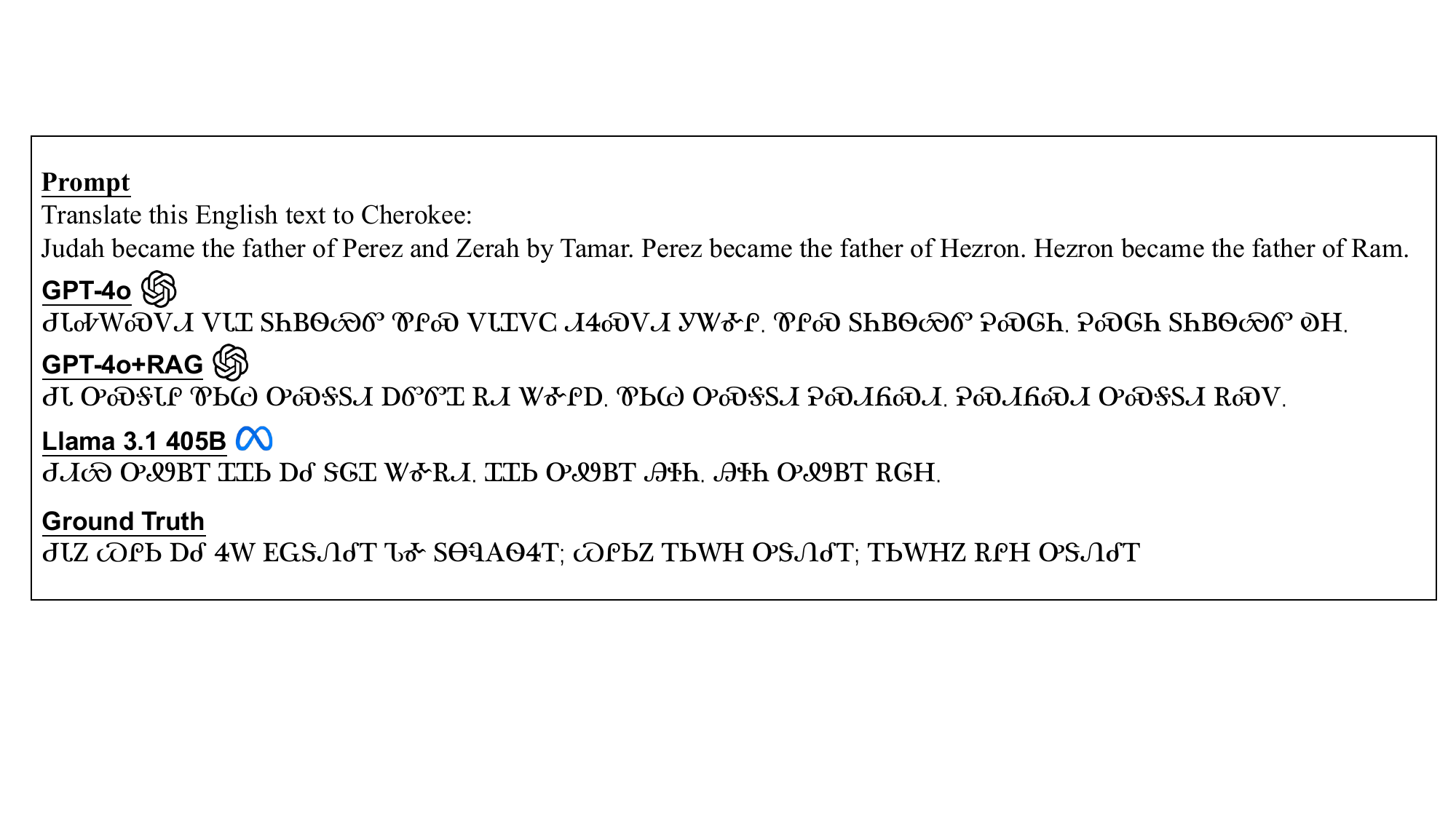}
  \caption{Additional testing example on Cherokee New Testament.}
  \label{fig:3}
\end{figure}

\begin{figure}[H]
  \centering
  \includegraphics[trim=20 100 10 80,clip,width=\linewidth]{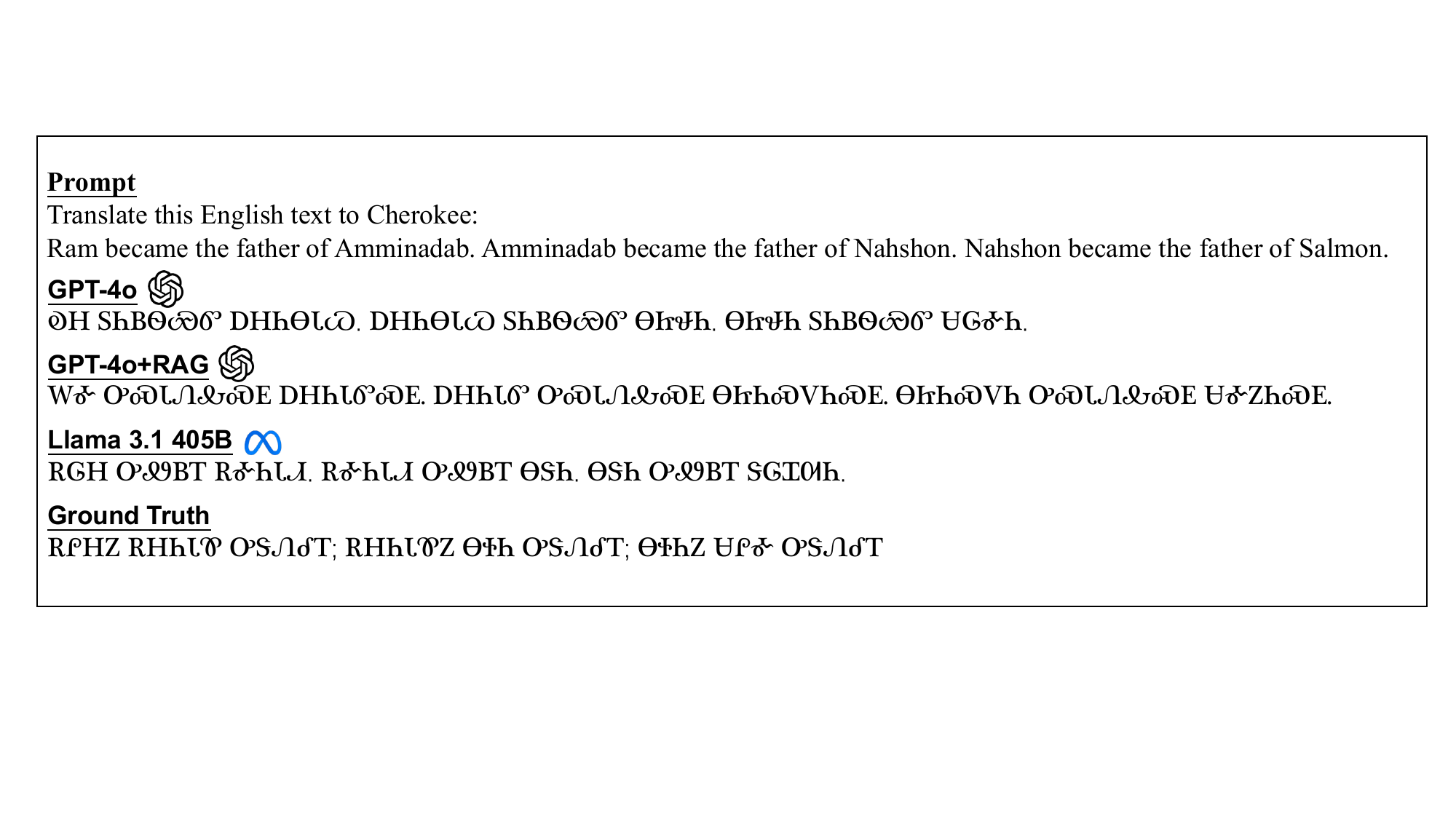}
  \caption{Additional testing example on Cherokee New Testament.}
  \label{fig:4}
\end{figure}

\begin{figure}[H]
  \centering
  \includegraphics[trim=20 120 30 80,clip,width=\linewidth]{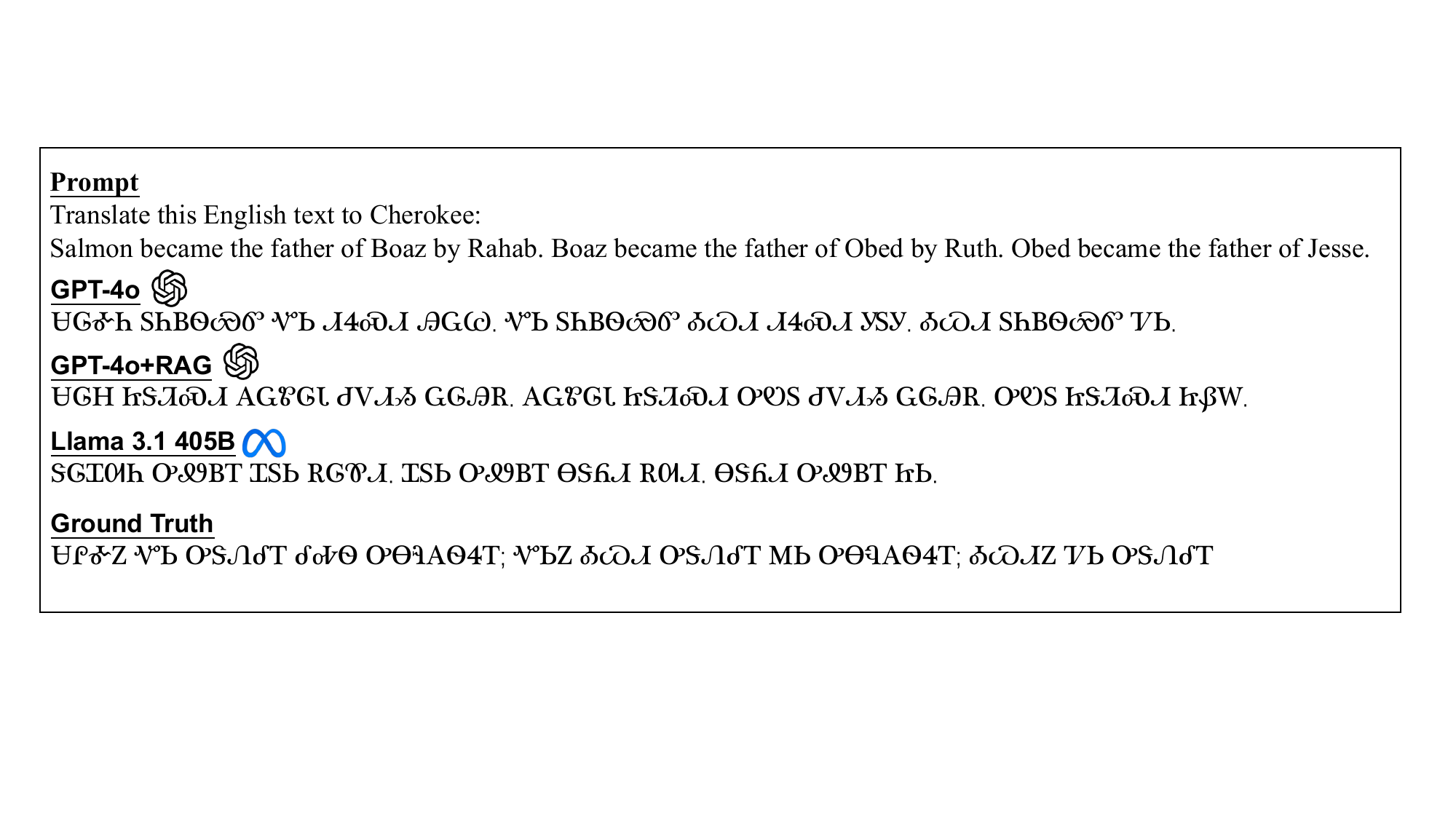}
  \caption{Additional testing example on Cherokee New Testament.}
  \label{fig:5}
\end{figure}

\begin{figure}[H]
  \centering
  \includegraphics[trim=40 120 38 80,clip,width=\linewidth]{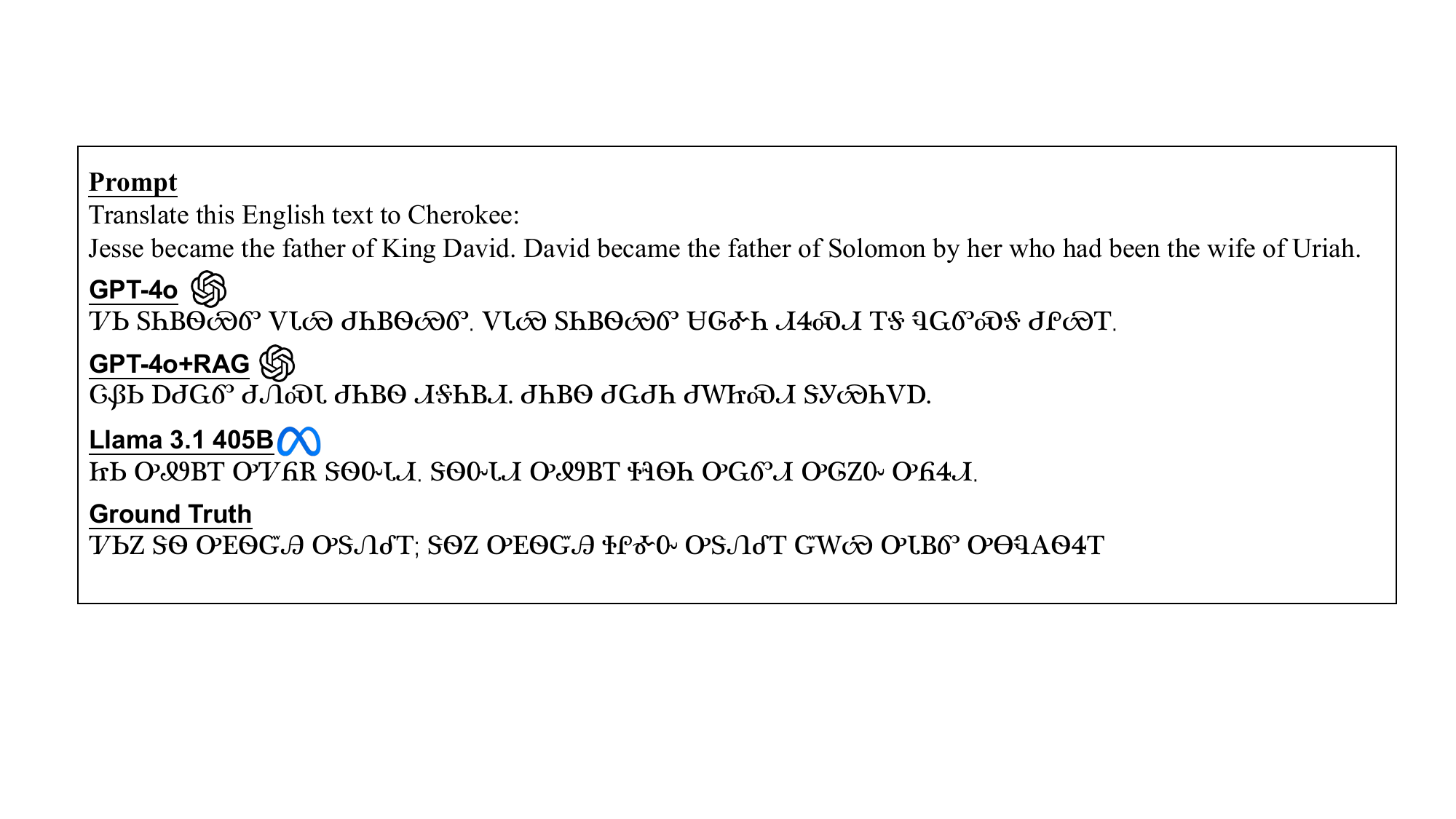}
  \caption{Additional testing example on Cherokee New Testament.}
  \label{fig:6}
\end{figure}

\begin{figure}[H]
  \centering
  \includegraphics[trim=30 120 70 80,clip,width=\linewidth]{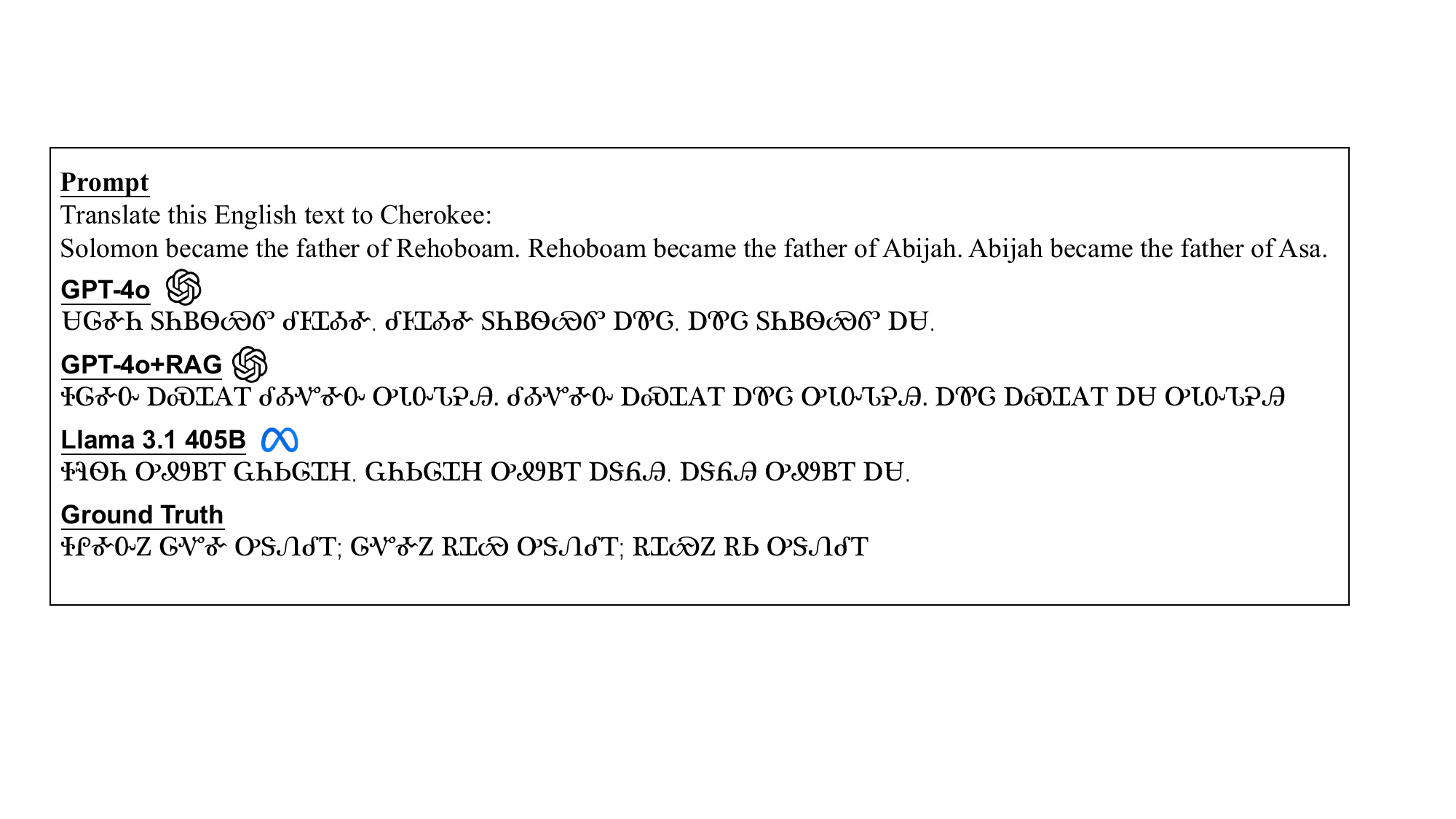}
  \caption{Additional testing example on Cherokee New Testament.}
  \label{fig:7}
\end{figure}

\begin{figure}[H]
  \centering
  \includegraphics[trim=36 100 55 80,clip,width=\linewidth]{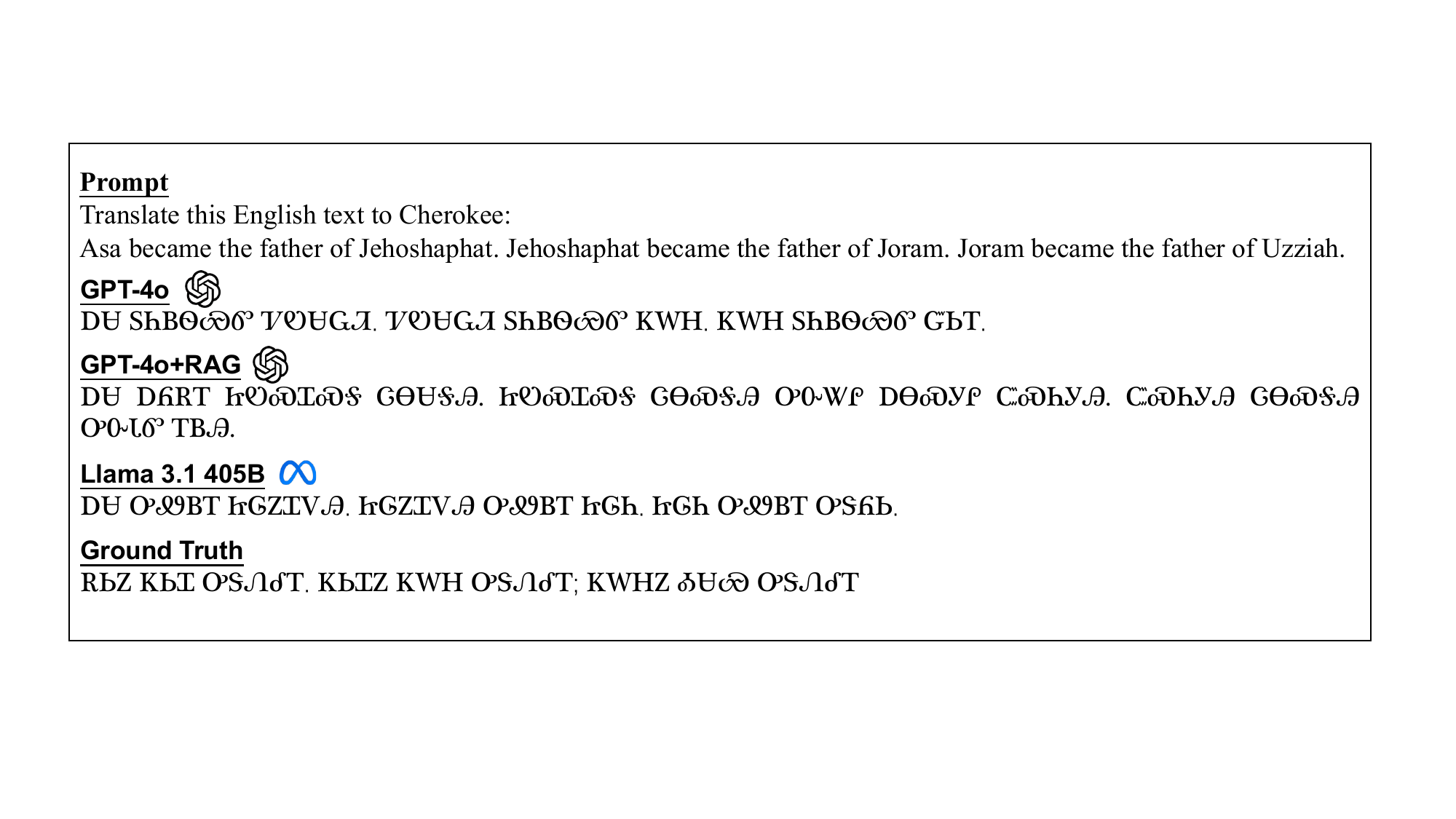}
  \caption{Additional testing example on Cherokee New Testament.}
  \label{fig:8}
\end{figure}

\begin{figure}[H]
  \centering
  \includegraphics[trim=68 100 58 80,clip,width=\linewidth]{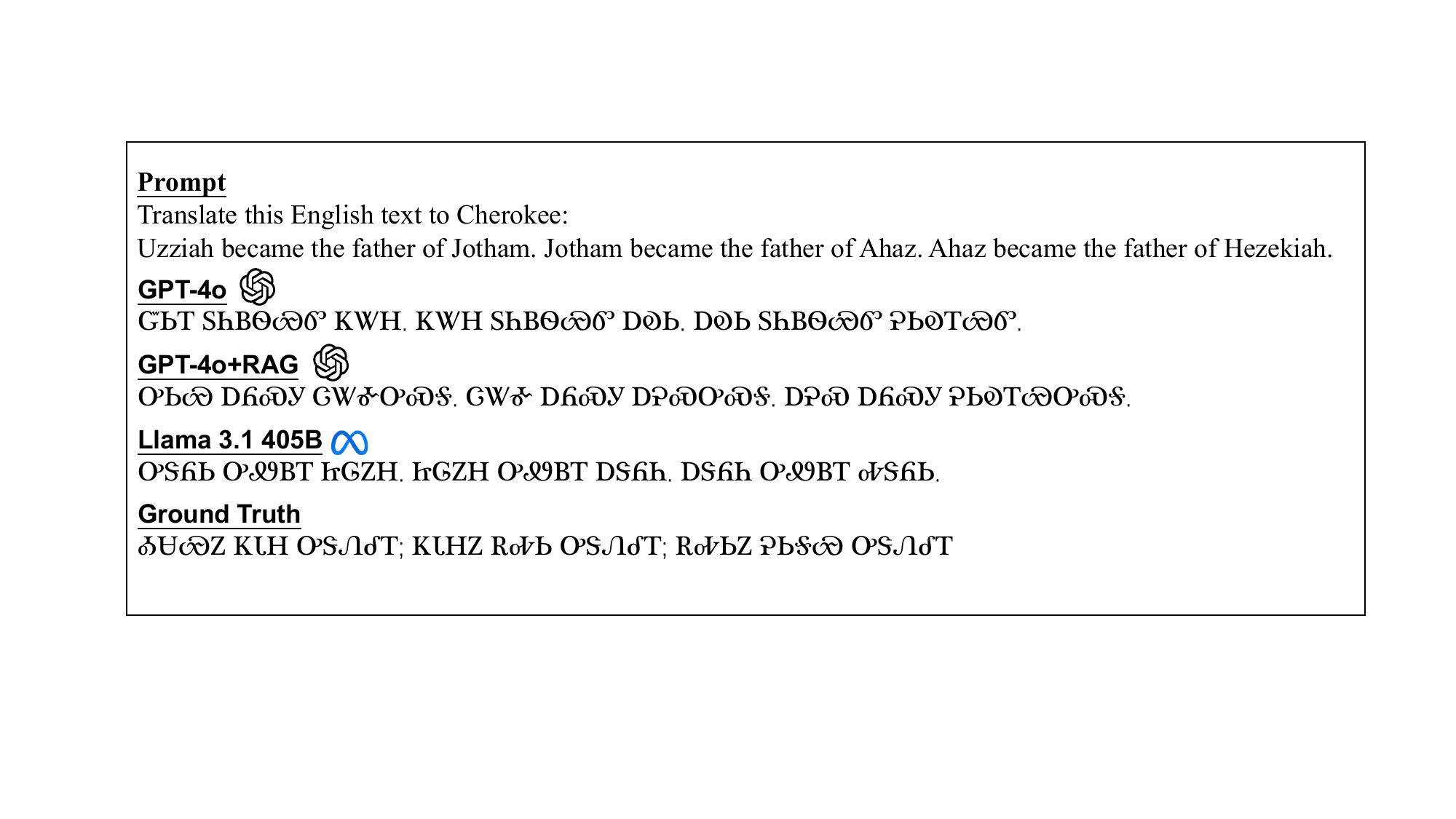}
  \caption{Additional testing example on Cherokee New Testament.}
  \label{fig:9}
\end{figure}

\begin{figure}[H]
  \centering
  \includegraphics[trim=68 100 31 80,clip,width=\linewidth]{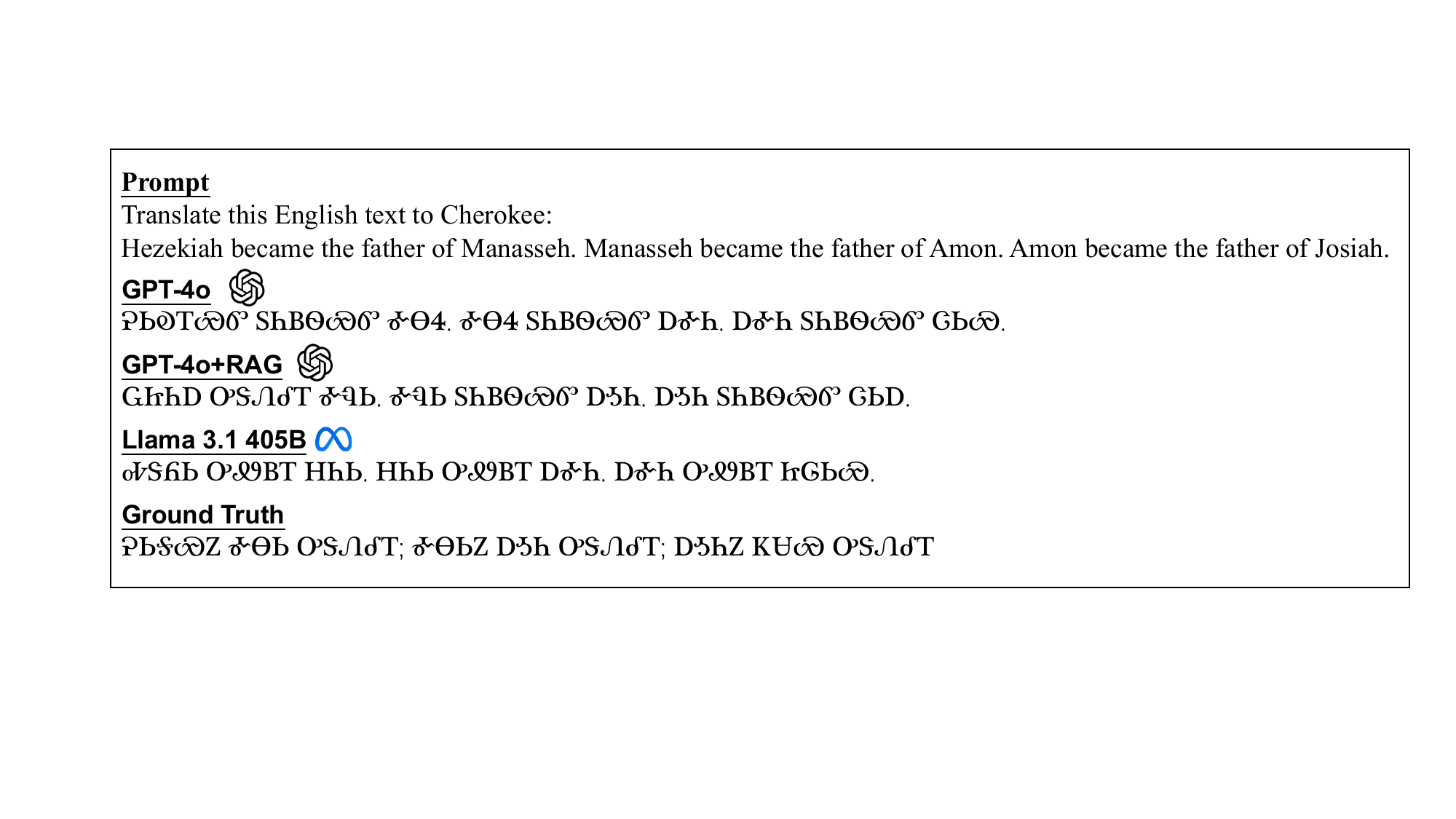}
  \caption{Additional testing example on Cherokee New Testament.}
  \label{fig:10}
\end{figure}

\begin{figure}[H]
  \centering
  \includegraphics[trim=50 10 81 10,clip,width=1.0\linewidth]{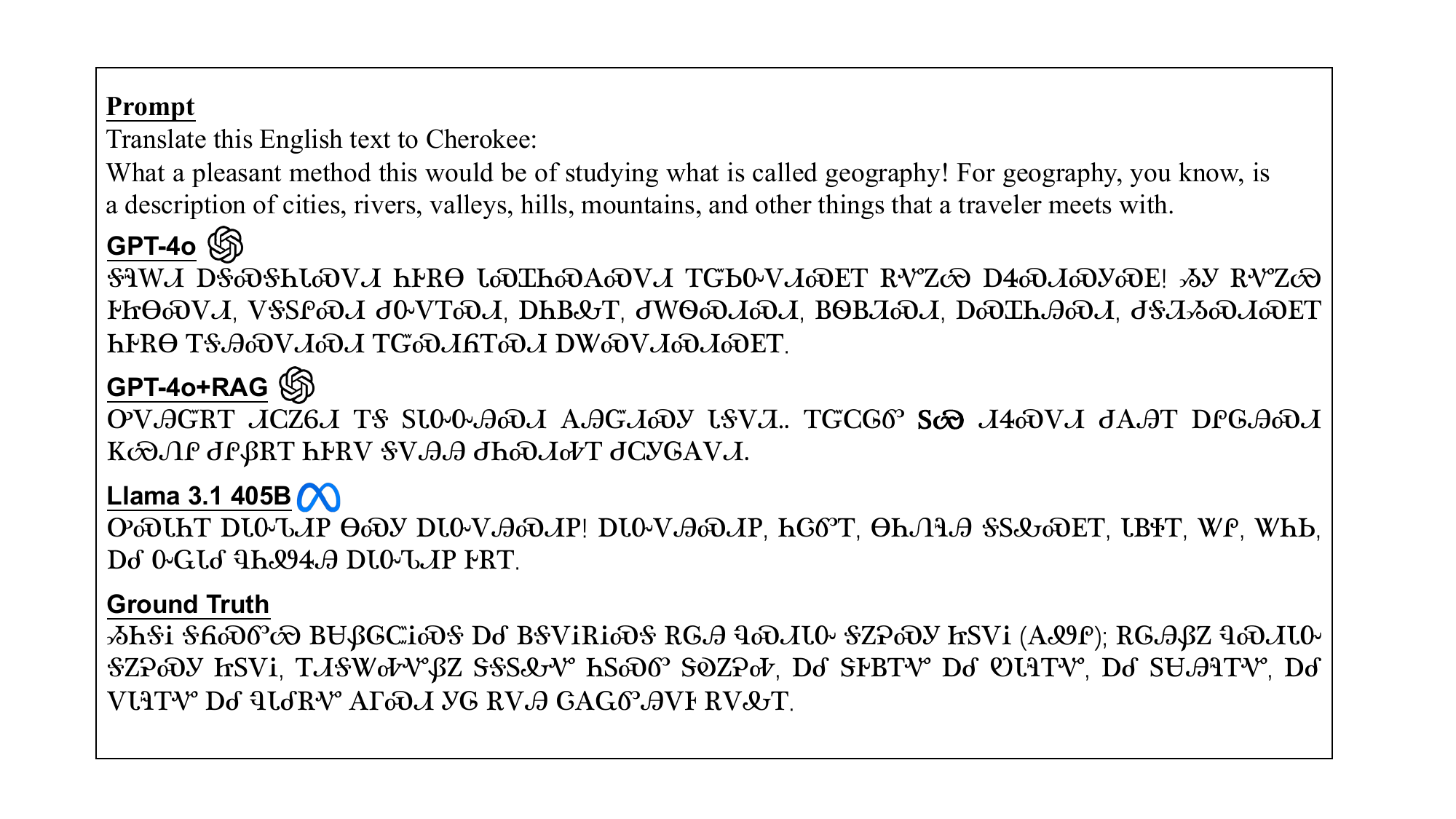}
  \caption{Additional testing example on Peter Parley’s Geography.}
  \label{fig:12}
\end{figure}

\begin{figure}[H]
  \centering
  \includegraphics[trim=30 0 40 10,clip,width=1.0\linewidth]{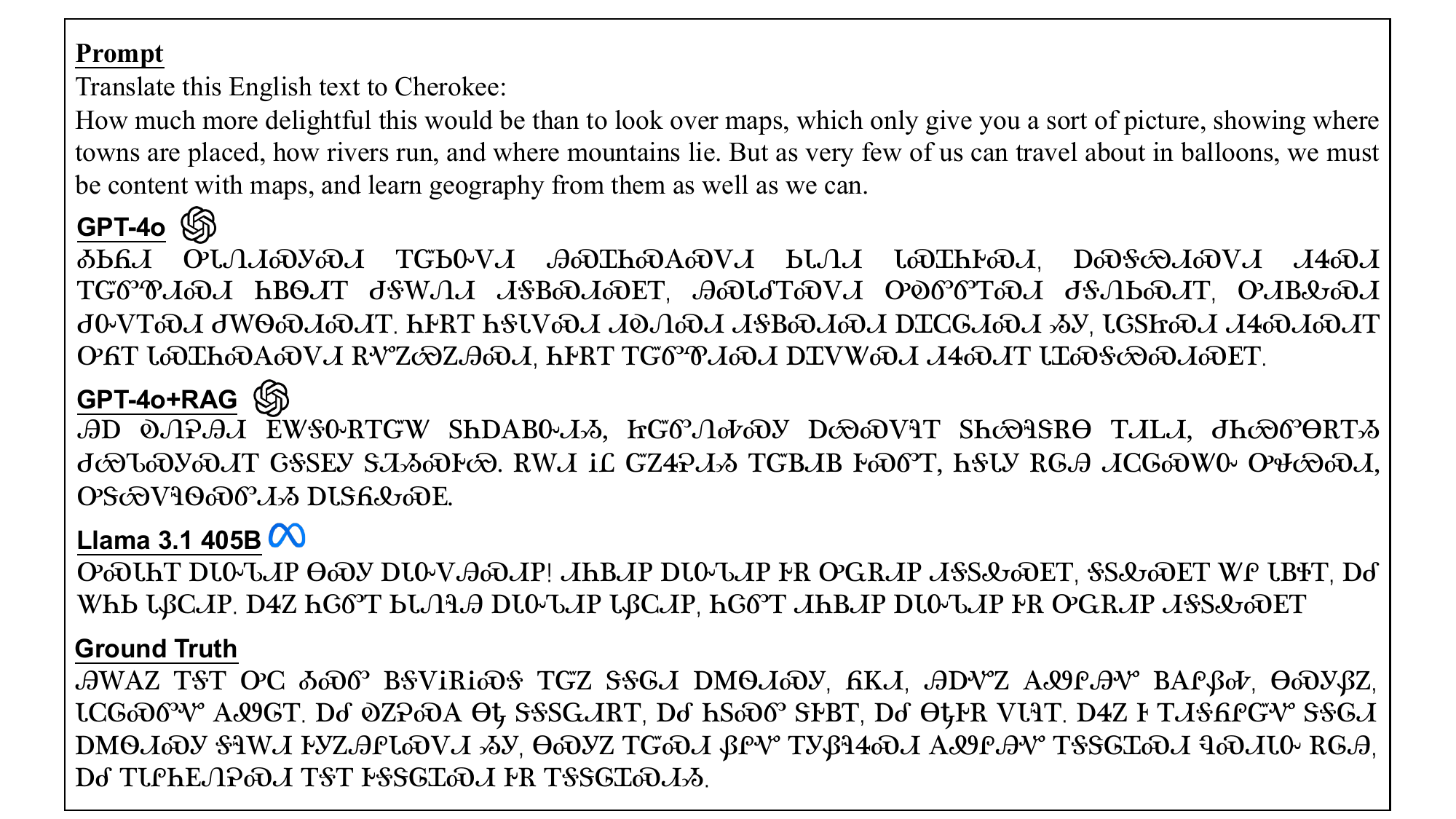}
  \caption{Additional testing example on Peter Parley’s Geography.}
  \label{fig:13}
\end{figure}

\begin{figure}[H]
  \centering
  \includegraphics[trim=30 0 32 10,clip,width=1.0\linewidth]{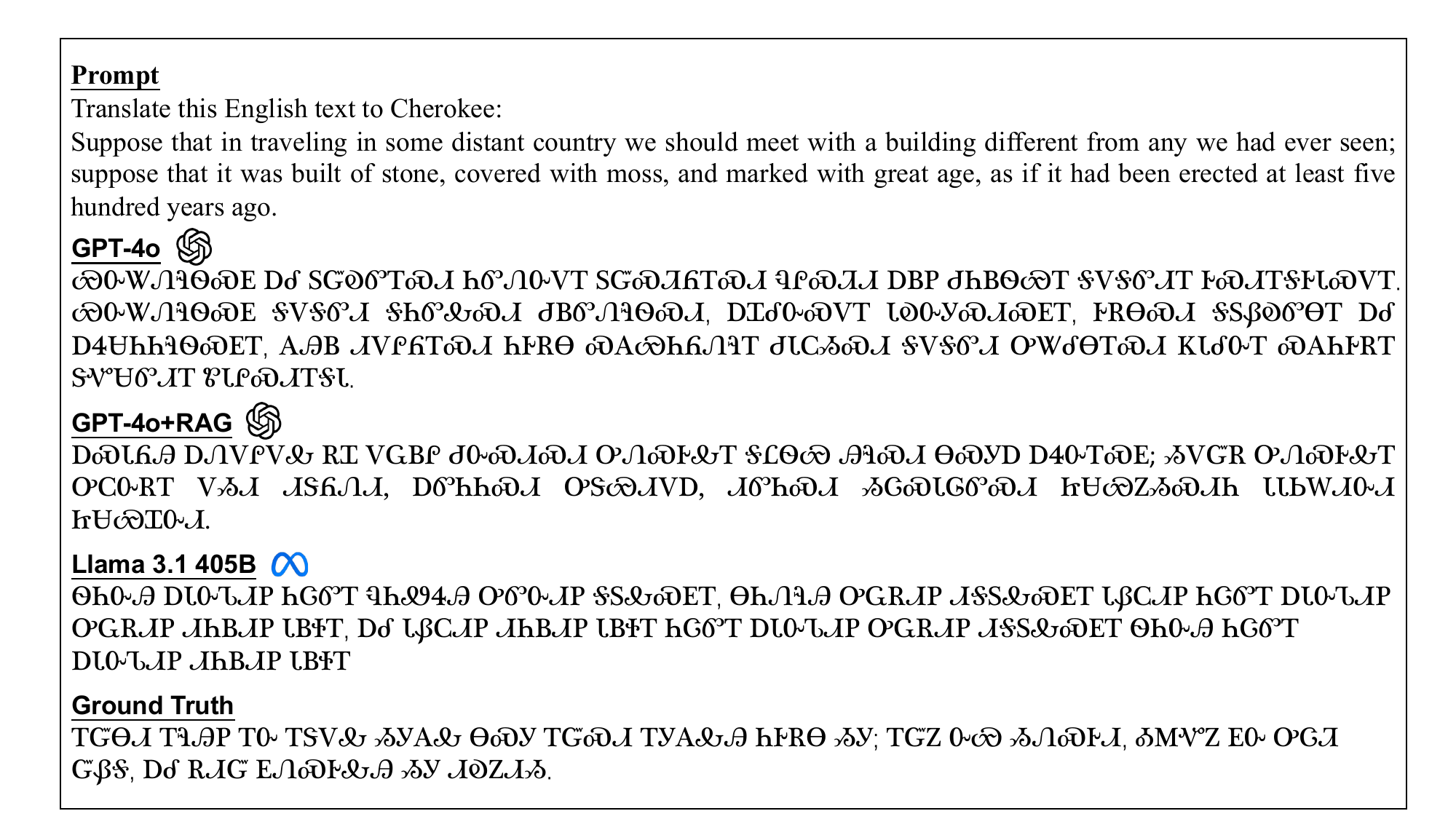}
  \caption{Additional testing example on Peter Parley’s Geography.}
  \label{fig:14}
\end{figure}

\begin{figure}[H]
  \centering
  \includegraphics[trim=20 0 40 10,clip,width=1.0\linewidth]{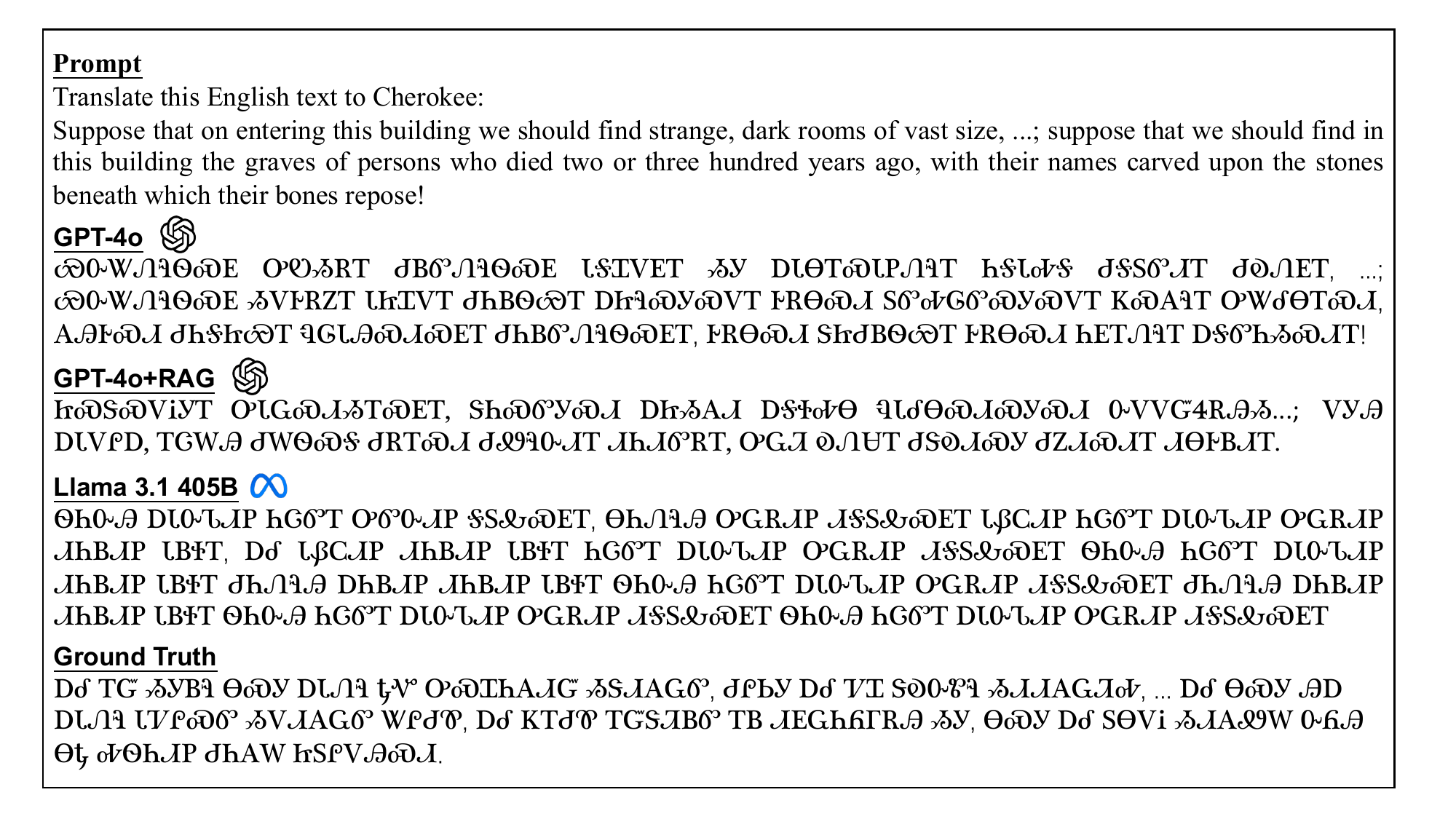}
  \caption{Additional testing example on Peter Parley’s Geography.}
  \label{fig:15}
\end{figure}

\begin{figure}[H]
  \centering
  \includegraphics[trim=20 0 10 10,clip,width=1.0\linewidth]{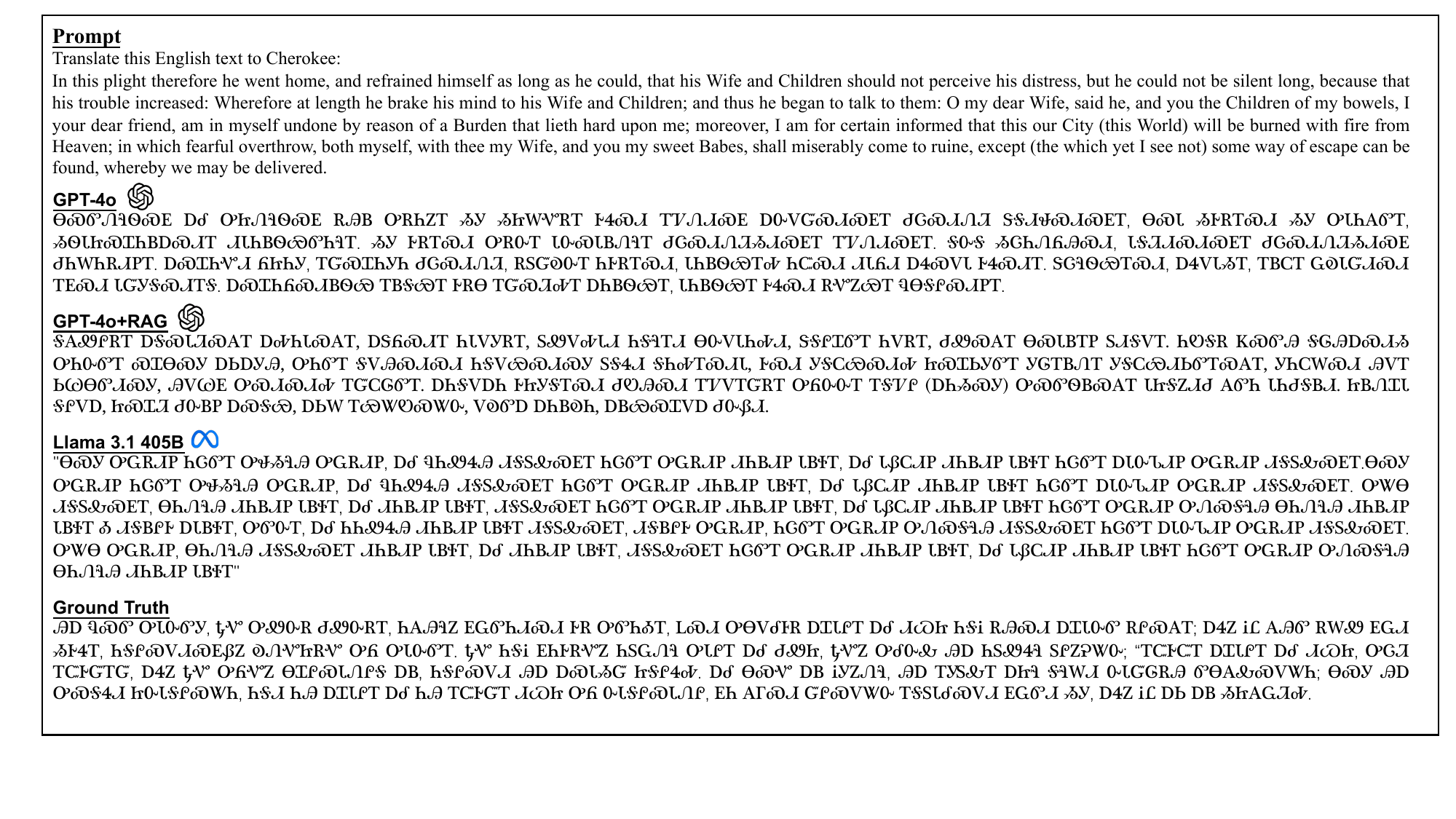}
  \caption{Additional testing example on The Pilgrim’s Progress.}
  \label{fig:17}
\end{figure}

\begin{figure}[H]
  \centering
  \includegraphics[trim=20 0 10 9,clip,width=1.0\linewidth]{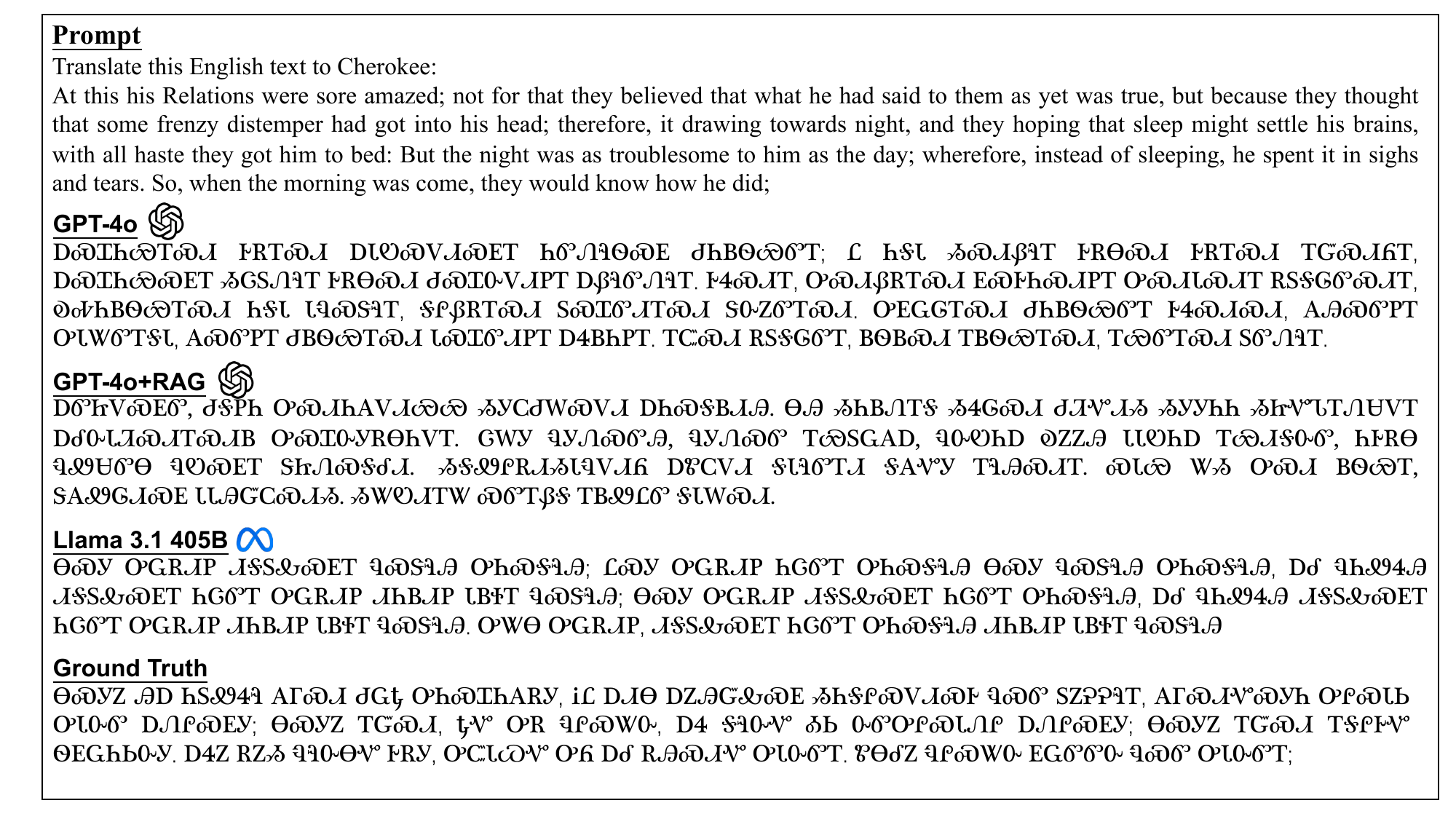}
  \caption{Additional testing example on The Pilgrim’s Progress.}
  \label{fig:18}
\end{figure}

\begin{figure}[H]
  \centering
  \includegraphics[trim=20 0 10 9,clip,width=1.0\linewidth]{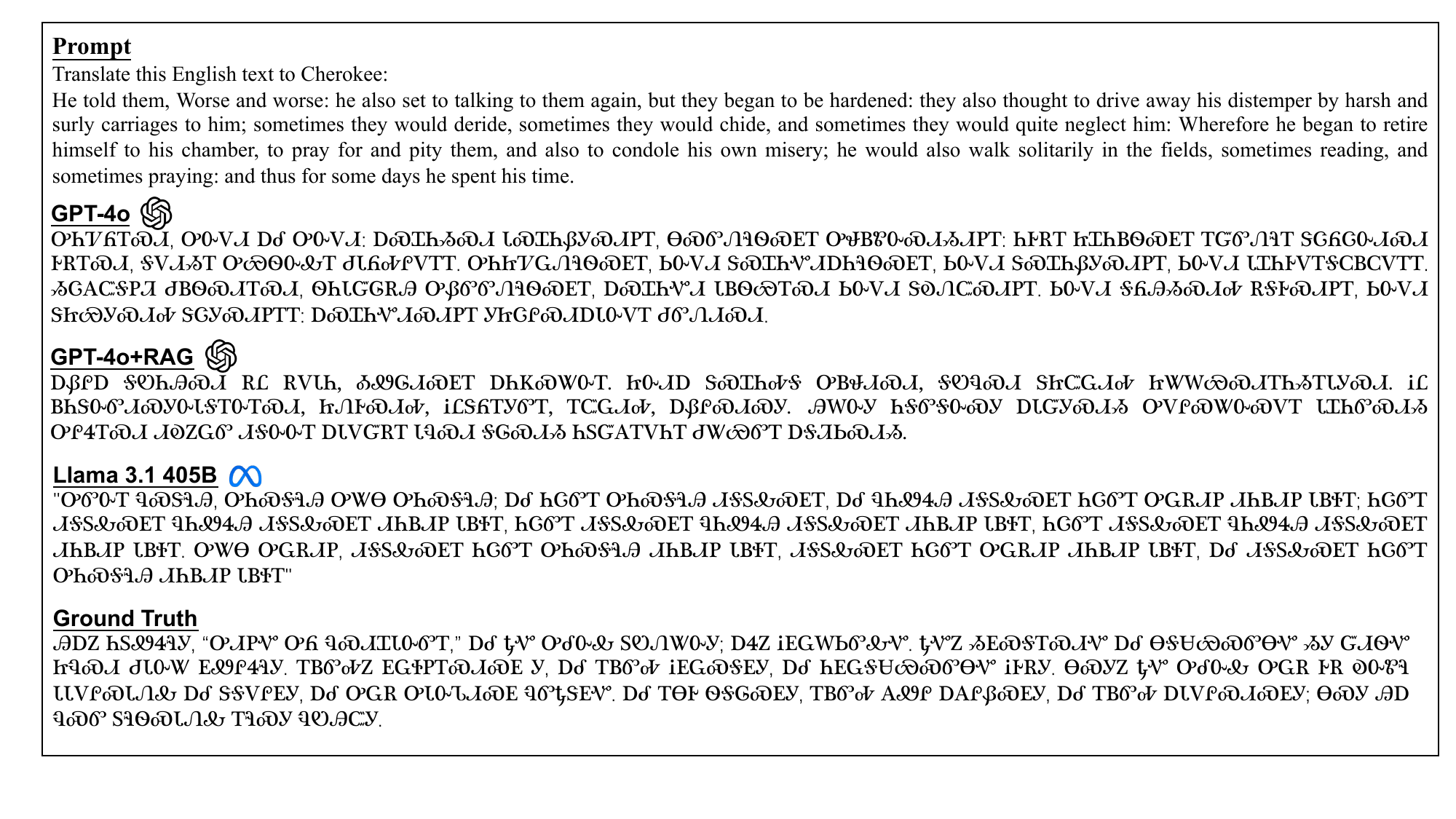}
  \caption{Additional testing example on The Pilgrim’s Progress.}
  \label{fig:19}
\end{figure}

\begin{figure}[H]
  \centering
  \includegraphics[trim=18 0 10 9,clip,width=1.0\linewidth]{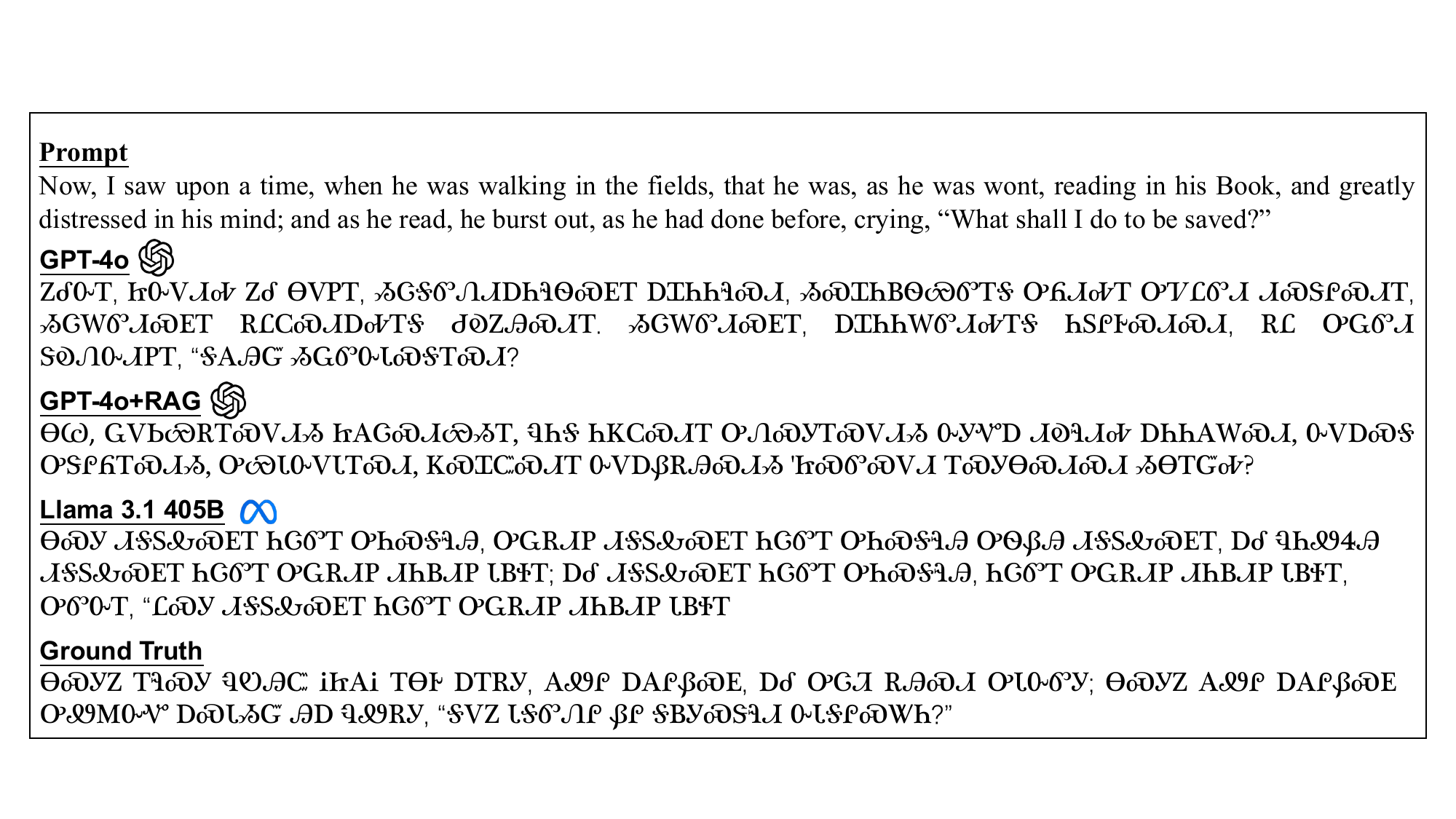}
  \caption{Additional testing example on The Pilgrim’s Progress.}
  \label{fig:20}
\end{figure}

\begin{figure}[H]
  \centering
  \includegraphics[trim=90 80 30 80,clip,width=\linewidth]{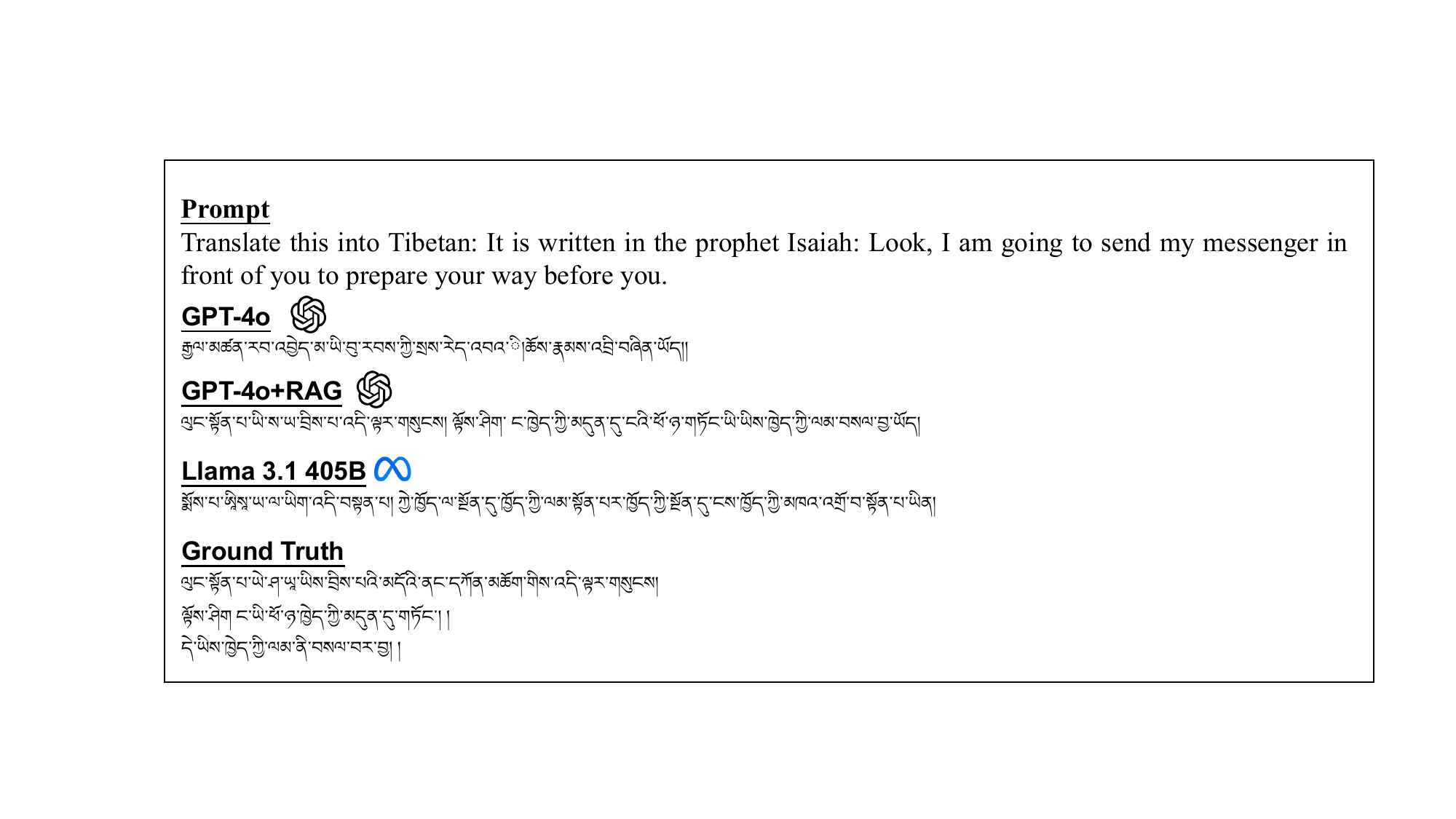}
  \caption{Additional testing example for Tibetan translation.}
  \label{fig:22}
\end{figure}

\begin{figure}[H]
  \centering
  \includegraphics[trim=30 40 30 100,clip,width=\linewidth]{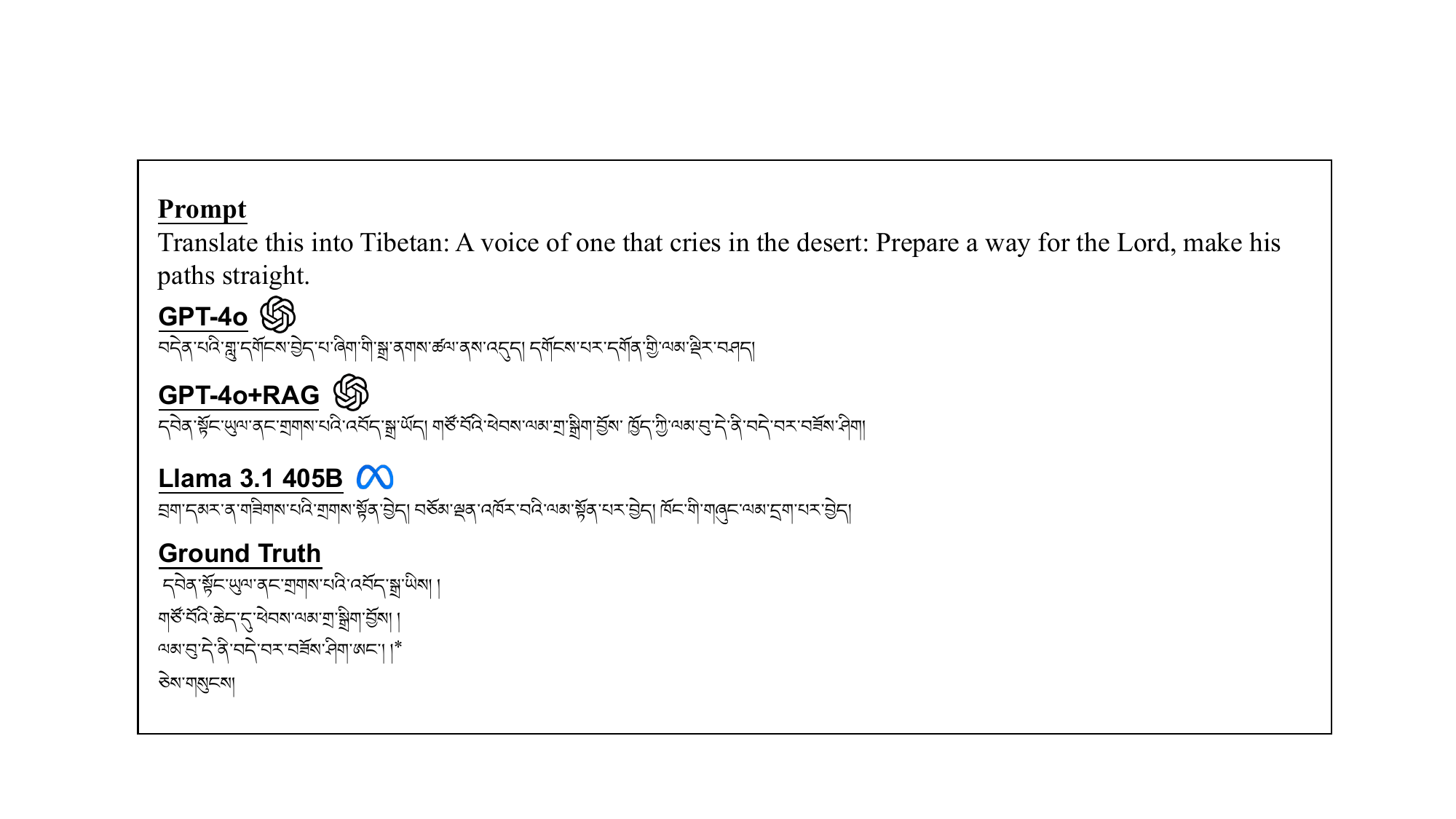}
  \caption{Additional testing example for Tibetan translation.}
  \label{fig:23}
\end{figure}

\begin{figure}[H]
  \centering
  \includegraphics[trim=30 80 30 100,clip,width=\linewidth]{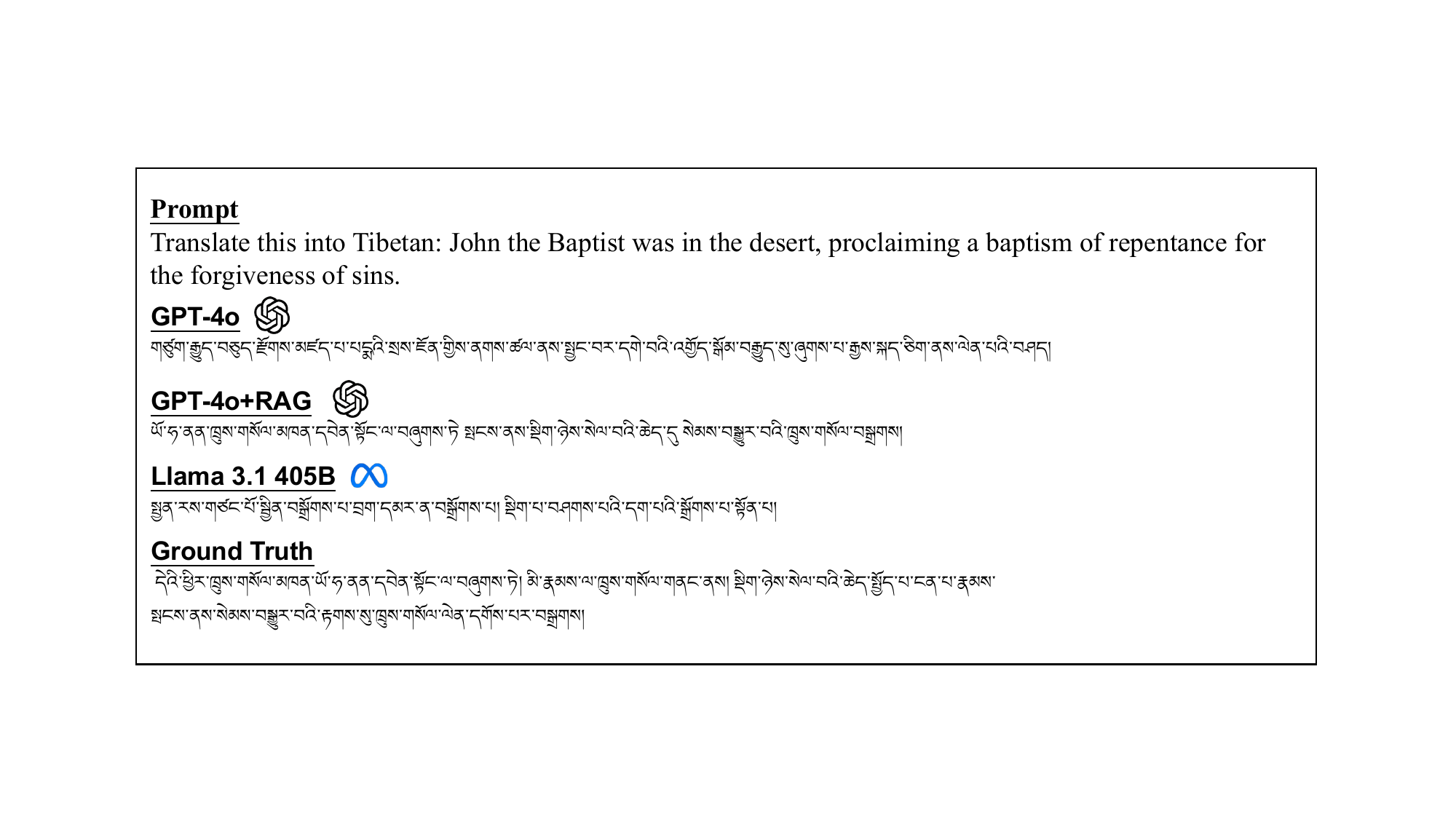}
  \caption{Additional testing example for Tibetan translation.}
  \label{fig:24}
\end{figure}

\begin{figure}[H]
  \centering
  \includegraphics[trim=50 80 30 100,clip,width=\linewidth]{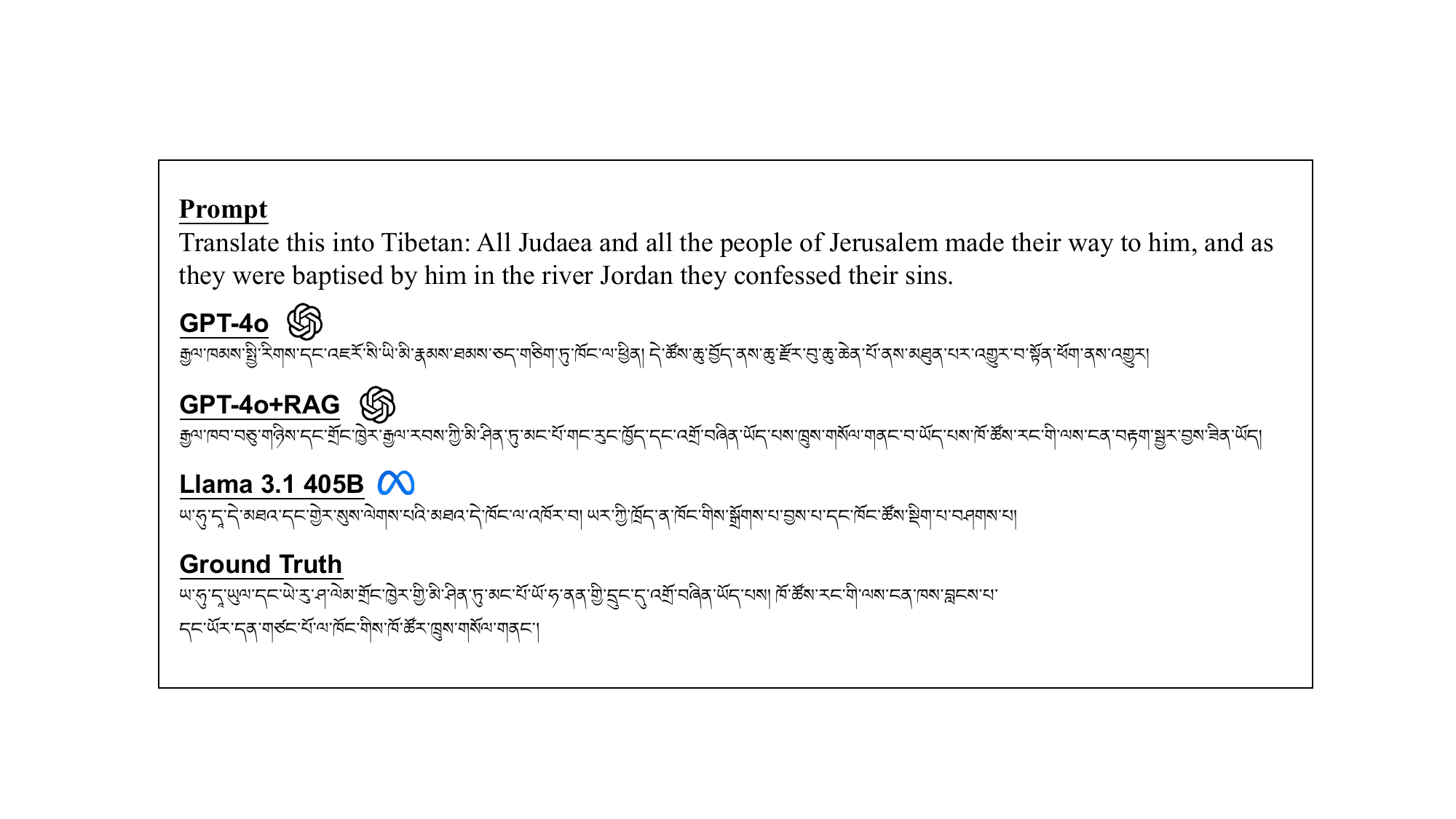}
  \caption{Additional testing example for Tibetan translation.}
  \label{fig:25}
\end{figure}

\begin{figure}[H]
  \centering
  \includegraphics[trim=50 50 100 100,clip,width=\linewidth]{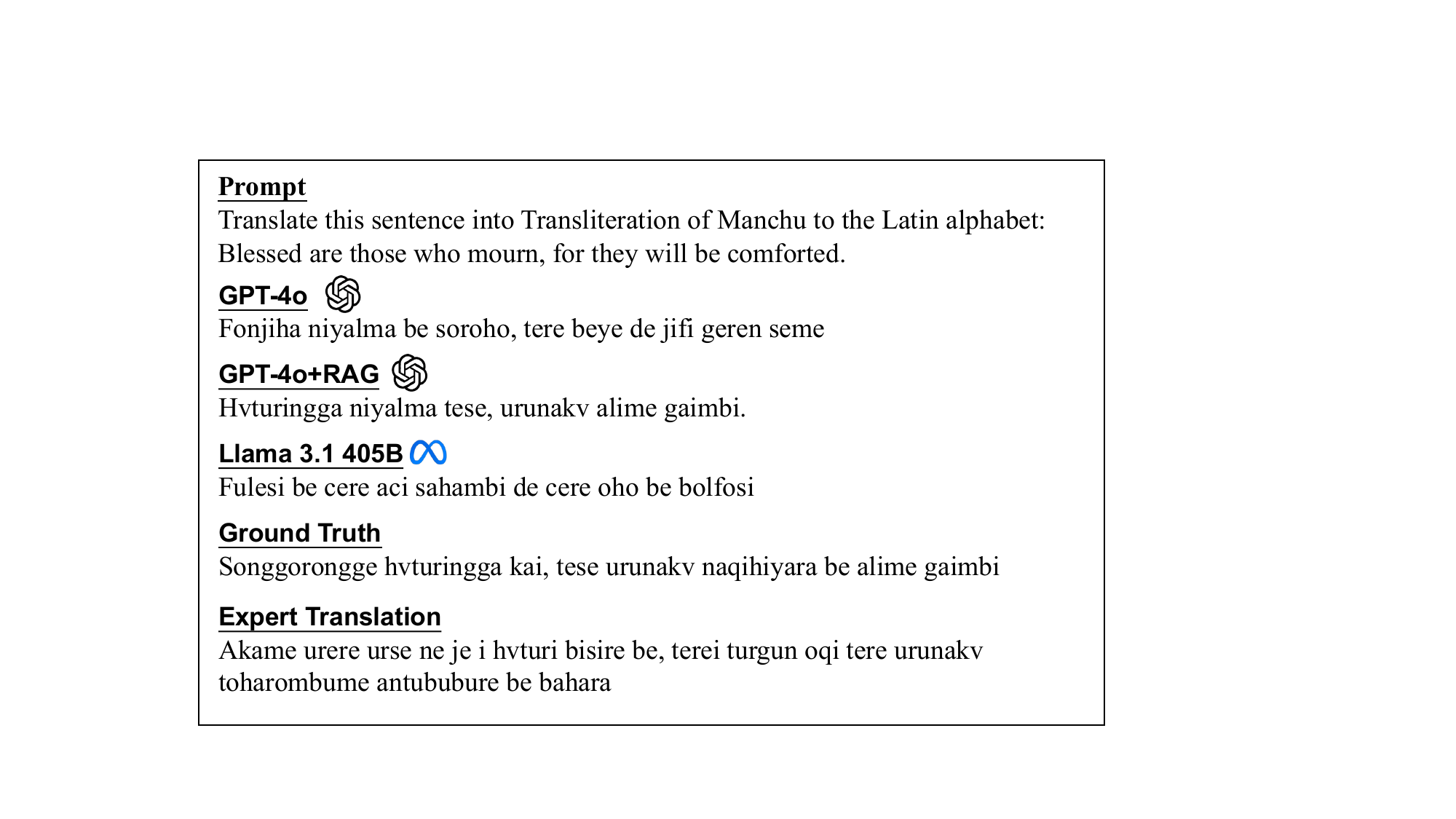}
  \caption{Additional testing example for Manchu translation.}
  \label{fig:27}
\end{figure}

\begin{figure}[H]
  \centering
  \includegraphics[trim=10 50 150 100,clip,width=\linewidth]{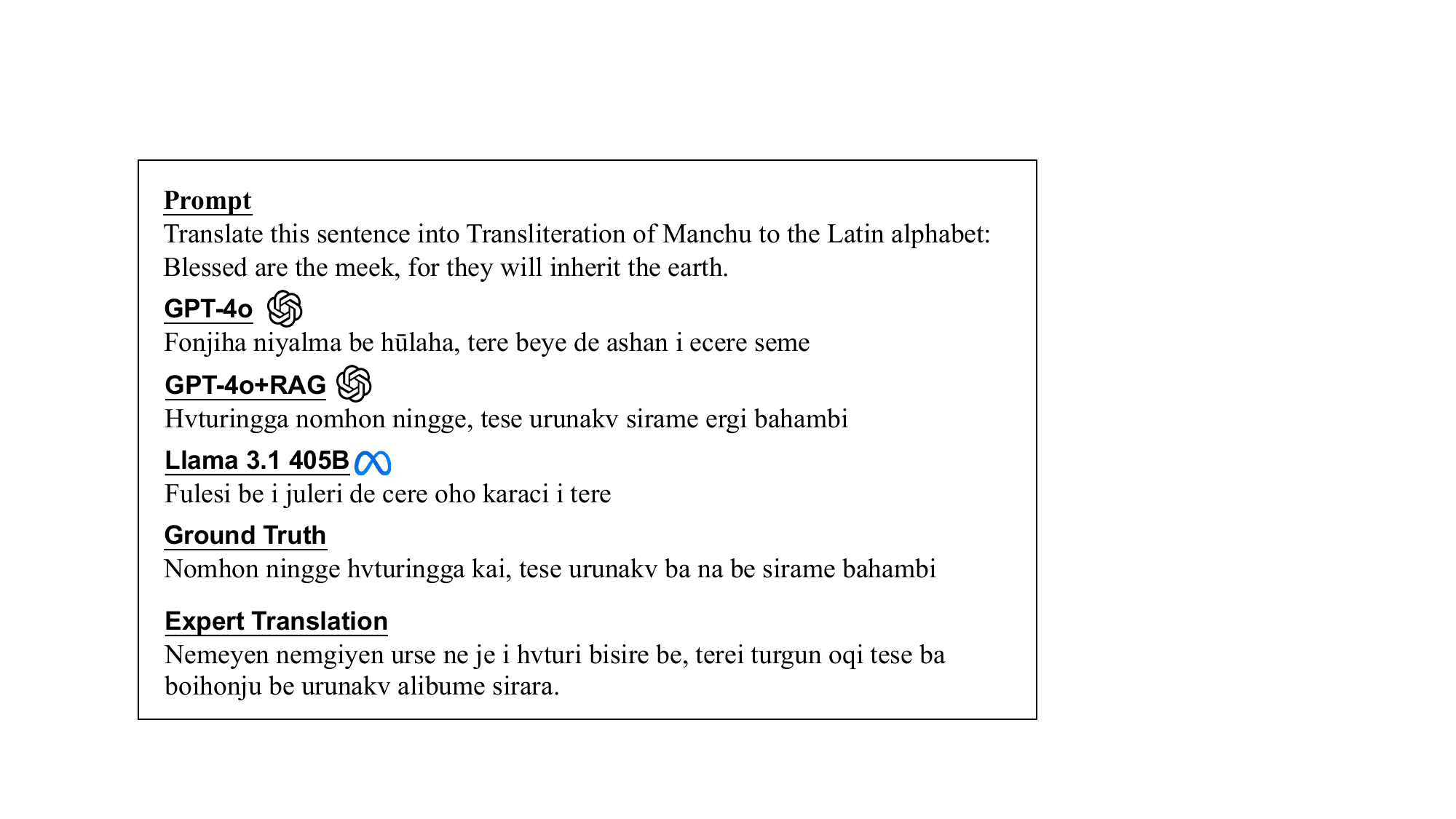}
  \caption{Additional testing example for Manchu translation.}
  \label{fig:28}
\end{figure}

\begin{figure}[H]
  \centering
  \includegraphics[trim=30 10 30 100,clip,width=\linewidth]{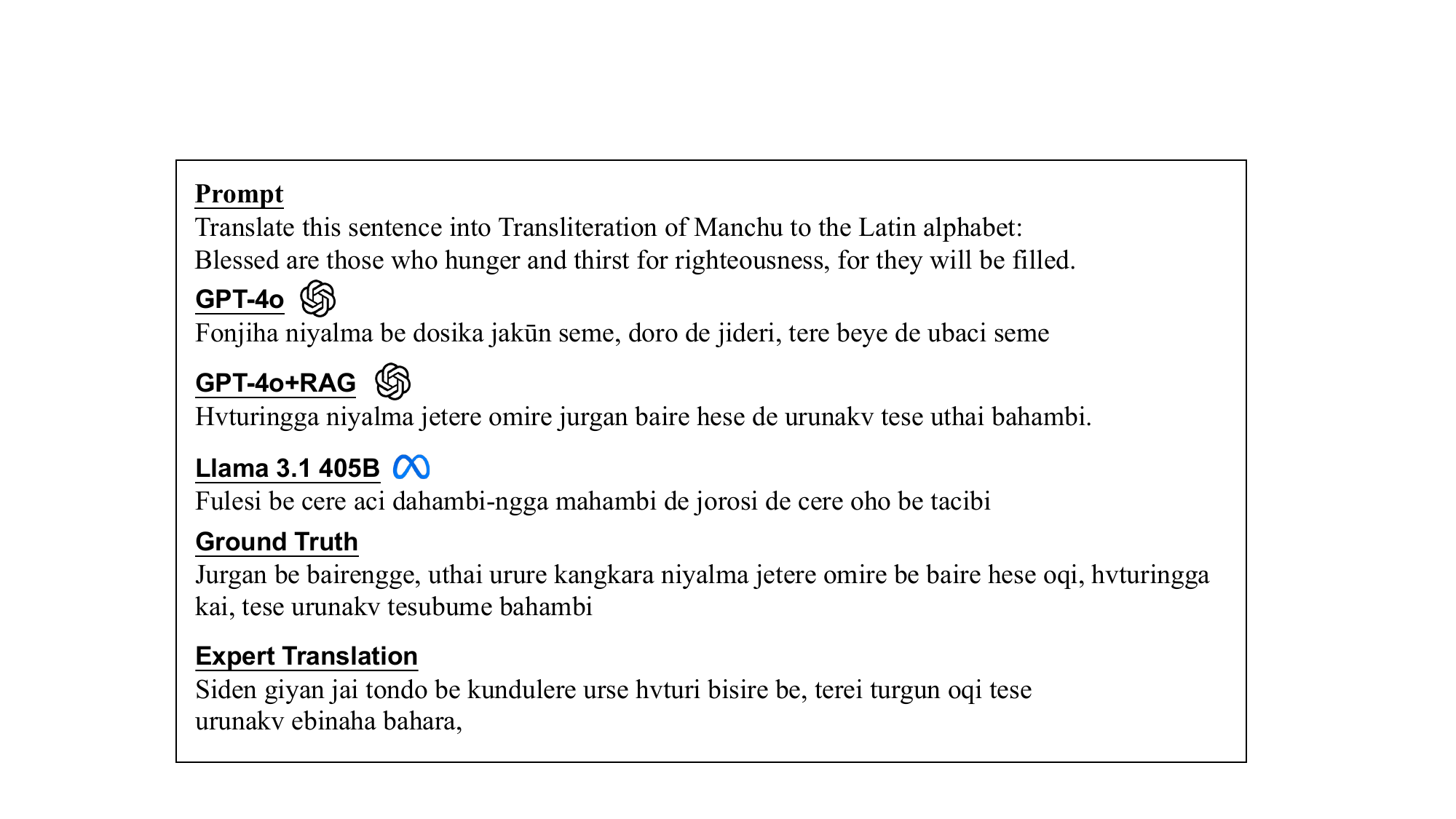}
  \caption{Additional testing example for Manchu translation.}
  \label{fig:29}
\end{figure}

\begin{figure}[H]
  \centering
  \includegraphics[trim=30 30 100 100,clip,width=\linewidth]{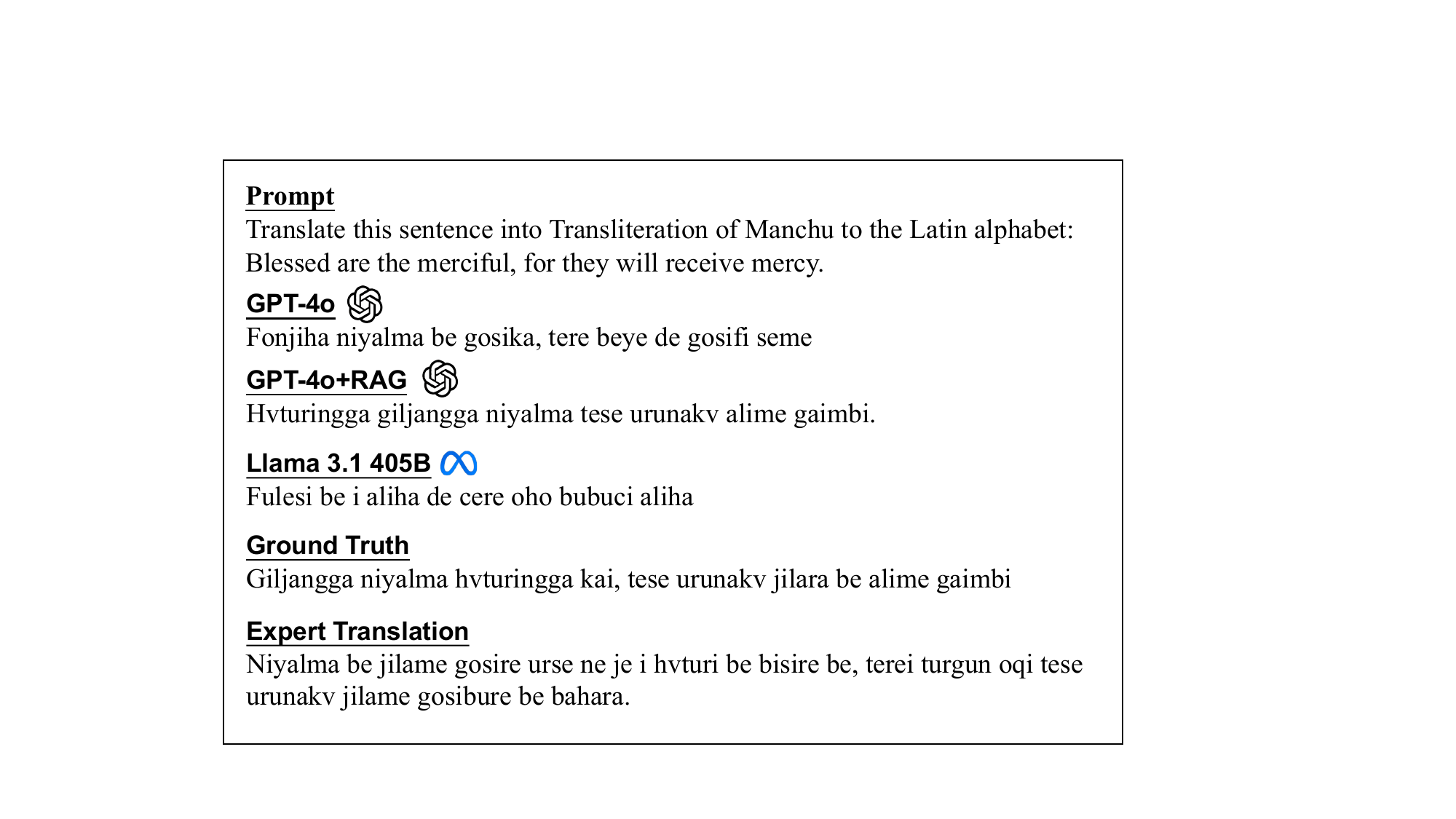}
  \caption{Additional testing example for Manchu translation.}
  \label{fig:30}
\end{figure}

\end{document}